\documentclass[10pt,twocolumn,letterpaper]{article}
\usepackage[accsupp]{axessibility}  

\usepackage{cvpr}              

\usepackage[dvipsnames]{xcolor}

\definecolor{cvprblue}{rgb}{0.21,0.49,0.74}
\usepackage[pagebackref,breaklinks,colorlinks,allcolors=cvprblue]{hyperref}
\usepackage{threeparttable}
\usepackage{booktabs}
\usepackage{amsopn}
\usepackage{multirow}
\usepackage{colortbl}
\usepackage{makecell}
\usepackage{float}

\usepackage{tabularx}
\usepackage{bbding}
\usepackage{pifont}

\usepackage{amsmath} 
\usepackage{amssymb}
\usepackage{amsfonts}
\usepackage{bm}
\usepackage{array}
\usepackage{dsfont}

\usepackage{wrapfig}
\usepackage{overpic}

\def\confName{CVPR}
\def\confYear{2025}

\newcommand*{\affaddr}[1]{#1} 
\newcommand*{\affmark}[1][*]{\textsuperscript{#1}}

\makeatletter
\def\blfootnote{\xdef\@thefnmark{}\@footnotetext}
\makeatother
\title{Towards Open-Vocabulary Audio-Visual Event Localization}

\author{Jinxing Zhou\affmark[1] \qquad Dan Guo\affmark[2] \qquad Ruohao Guo\affmark[3] \qquad Yuxin Mao\affmark[4] \qquad Jingjing Hu\affmark[2]\\Yiran Zhong\affmark[5] \qquad Xiaojun Chang\affmark[1,6] \qquad Meng Wang\affmark[2,*]\\[5pt]
\affaddr{\affmark[1]MBZUAI}\quad
\affaddr{\affmark[2]HFUT}\quad
\affaddr{\affmark[3]PKU} \quad \affaddr{\affmark[4]NWPU} \quad
\affaddr{\affmark[5]OpenNLPLab} \quad\affaddr{\affmark[6]USTC}\\
\texttt{\url{https://github.com/jasongief/OV-AVEL}}
}

\begin{document}
\maketitle

\blfootnote{$^{*}$: Meng Wang is the corresponding author (wangmeng@hfut.edu.cn)}

\begin{abstract}
The Audio-Visual Event Localization (AVEL) task aims to temporally locate and classify video events that are both audible and visible.
Most research in this field assumes a closed-set setting, which restricts these models' ability to handle test data containing event categories absent (unseen) during training. 
Recently, a few studies have explored AVEL in an open-set setting, enabling the recognition of unseen events as ``unknown'', but without providing category-specific semantics.
In this paper, we advance the field by introducing the Open-Vocabulary Audio-Visual Event Localization (OV-AVEL) problem, which requires localizing audio-visual events and predicting explicit categories for both seen and unseen test data at inference.
To address this new task, we propose the OV-AVEBench dataset, comprising 24,800 videos across 67 real-life audio-visual scenes (seen:unseen = 46:21), each with manual segment-level annotation.
We also establish three evaluation metrics for this task.
Moreover, we investigate two baseline approaches, one training-free and one using a further fine-tuning paradigm.
Specifically, we utilize the unified multimodal space from the pretrained ImageBind model to extract audio, visual, and textual (event classes) features.
The training-free baseline then determines predictions by comparing the consistency of audio-text and visual-text feature similarities.
The fine-tuning baseline incorporates lightweight temporal layers to encode temporal relations within the audio and visual modalities, using OV-AVEBench training data for model fine-tuning.
We evaluate these baselines on the proposed OV-AVEBench dataset and discuss potential directions for future work in this new field.
\end{abstract}

\begin{figure}[t]
  \centering
\includegraphics[width=0.48\textwidth]{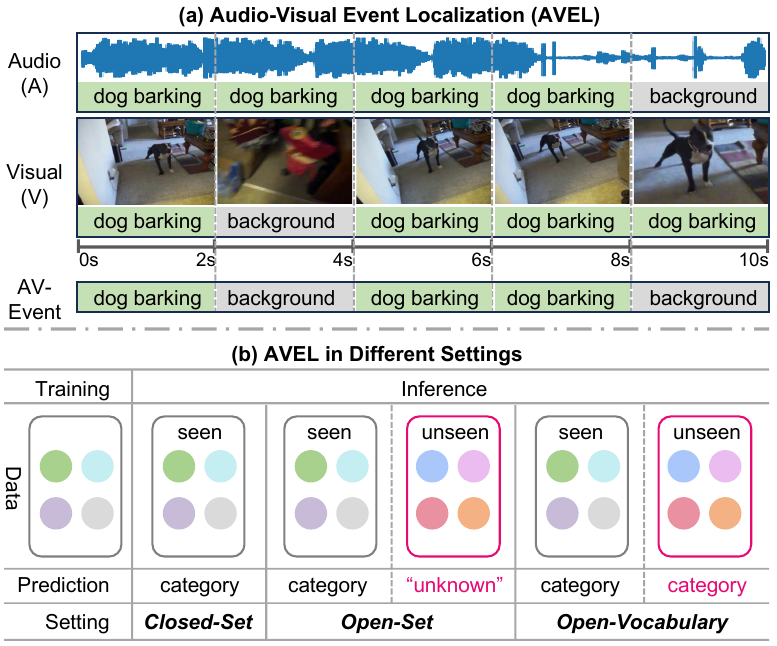}
\vspace{-5ex}
   \caption{{ \textbf{(a) Illustration of the AVEL task}, which aims to temporally localize segments containing events that are both audible and visible, and identify their categories. \textbf{(b) Studies of AVEL in different settings.} In contrast to previous closed-set and open-set settings, we explore a more practical open-vocabulary AVEL problem, which needs to infer explicit event categories for both seen and unseen test data (\ie, data containing classes seen and unseen during training). Each color represents a distinct event class.} 
   }
   \vspace{-3ex}
   \label{fig:introduction}
\end{figure}

\section{Introduction}\label{sec:intro}
Audio-visual learning, an essential sub-field of multimodal learning, has garnered increasing attention in recent years, resulting in the development of various research tasks, such as 
sound-source localization ~\cite{qian2020multiple, chen2021localizing,guo2024unitr,guo2024instance,mo2022localizing,mahmud2024t} and segmentation~\cite{zhou2022avs,zhou2023avss,hao2024improving,mao2023contrastive,mao2023multimodal,guo2023avis,guo2024open}, audio-visual video parsing~\cite{tian2020unified,yu2021mm,gao2023collecting,zhou2023improving,zhou2024label,zhou2024vaplan,zhao2024multimodal,chen2024cm} and generation~\cite{shen2023fine,mao2024tavgbench,li2024tri}, question answering~\cite{li2022learning,yang2022avqa,li2023progressive,li2024object,li2024patch}, \etc.
In this paper, we focus on a fundamental research task of Audio-Visual Event Localization (AVEL)~\cite{tian2018audio}.
As shown in Fig.~\ref{fig:introduction}(a), given a video containing both audio and visual streams, the AVEL task aims to temporally localize segments that contain an audio-visual event (\ie, both audible and visible) and identify its category. For segments that do not satisfy this condition (\ie, only audible/visible or neither), their category is assigned to a special \textit{background} class.
In other words, this task requires perceiving the semantic alignment between audio and visual modalities, known as audio-visual correspondence~\cite{arandjelovic2017look,arandjelovic2018objects}.

In recent years, there has been rapid advancement in the AVEL field:
{\textit{1) Closed-Set Audio-Visual Event Localization.}}
Since the pioneering work~\cite{tian2018audio}, numerous significant research works have been proposed. 
For example, these methods aim to improve audio-visual fusion~\cite{tian2018audio,wu2019dual,xu2020cross,lin2019dual,duan2021audio,zhou2024dense,he2024cace,zhou2021positive}, better distinguish the {background}~\cite{xia2022cross,zhou2022cpsp}, and localize more precise temporal boundaries~\cite{yu2021mm,wu2022span,mahmud2023ave}.
While these methods achieve satisfactory performance in traditional AVEL tasks, they are designed for a closed-set scenario.
As shown in Fig.~\ref{fig:introduction}(b), methods in this setting can only infer data with event \textit{classes} encountered or \textit{seen} during model training (referred to as \textbf{\textit{seen test data}} in our work), making it hard for \textbf{\textit{unseen test data}} (namely test data with \textit{classes unseen} during training) processing. \\
\noindent{\textit{2) Open-Set Audio-Visual Event Localization.}} 
Recently, Yu~\etal began exploring the AVEL task in an open-set setting~\cite{yu2024openave}.
To the best of our knowledge, this is currently the only work in this setting.
Specifically, the open-set AVEL needs to handle both seen and unseen test data at inference.
For the {unseen test data} with novel classes unseen in training, the model should recognize it as ``\textit{unknown}'' rather than classifying it into a known category.
By employing evidential deep learning and positive-unlabeled learning, ~\cite{yu2024openave} can identify unknown events in unseen test data.
However, the model remains unable to determine specific categories for unseen events.
Additionally, its model evaluation is conducted on a limited subset of the relatively small AVE~\cite{tian2018audio} dataset, where only 7 classes are treated as unknown, limiting its applicability in real-world scenarios.

In this paper, we investigate the {\textbf{Open-Vocabulary Audio-Visual Event Localization (OV-AVEL)}} problem, a novel and more practical extension of AVEL.
As shown in Fig.~\ref{fig:introduction}(b), OV-AVEL seeks to predict explicit event categories for both seen and unseen test data during inference, instead of assigning a general \textit{unknown} class to unseen data as in open-set AVEL, thus providing more detailed temporal localization results.
Notably, event categories in unseen test data are not present during model training.
A related topic to OV-AVEL is \textit{Audio-Visual Zero-Shot Learning (AV-ZSL)}~\cite{parida2020coordinated,mazumder2021avgzslnet,mercea2022audio,mercea2022temporal,mo2024audio}, which aims to classify unseen videos during testing by integrating both audio and visual signals.
The main difference is that AV-ZSL only needs to determine the category of the entire video, whereas our OV-AVEL seeks more fine-grained classification at the temporal level, requiring more precise recognition of audio-visual correspondence (\ie, perceiving the event category for each modality at each segment).

To support this new task, we develop a corresponding dataset named \textbf{{OV-AVEBench}}.
Compared to the AVE~\cite{tian2018audio} dataset used in closed-set and open-set AVEL, our OV-AVEBench offers a broader range of video and event categories.
An overall comparison is presented in Table~\ref{tab:dataset_comparison}.
Specifically, the proposed OV-AVEBench includes 24,800 videos in total, approximately 6 times the number in AVE~\cite{tian2018audio} dataset;
The videos in our OV-AVEBench encompass 67 classes of audio-visual events, whereas the AVE dataset includes only 28. 
Moreover, each video sample in OV-AVEBench is manually annotated at the segment level, providing precise labels for model training or fine-tuning.
These efforts in dataset construction allow us to explore more training data (seen classes) and unseen test data (unseen classes), facilitating model development and evaluation for real applications.
Details about data collection, annotation, and splitting will be presented in Sec.~\ref{sec:ov-avebench}.

In addition, we standardize the {\textbf{evaluation metrics}} for the studied OV-AVEL task.
Prior AVEL studies typically adopt \textit{accuracy}~\cite{tian2018audio} as the evaluation metric, which segment-wisely compares predictions and ground truths. This metric does not account for the \textit{recall} and may not be intuitive in evaluating predicted events across different temporal scales.
Inspired by the metrics in the audio-visual video parsing task~\cite{tian2020unified}, we incorporate the F1-score as an additional evaluation metric for OV-AVEL, measuring it at both the segment-level and event-level.
The segment-level F1-score is calculated by segment-wise comparison of predictions with ground truths.
Notably, the event-level metric treats consecutive segments with identical predictions as a complete event and computes the F1-score by assessing whether the Intersection over Union (IoU) between the predicted whole event and ground truth event exceeds the threshold of 0.5. Thus, this metric is stricter in evaluating the temporal boundaries of predictions.

\begin{table}[t]
    \centering
        \caption{\textbf{Comparison of our OV-AVEBench with other datasets used in various AVEL settings}.}
    \vspace{-2ex}
    \resizebox{\linewidth}{!}{
    \begin{tabular}{l|l|cc|ccc}
        \hline\toprule
        \multirow{2}{*}{Settings} & \multirow{2}{*}{Dataset} & \multicolumn{2}{c|}{Video} & \multicolumn{3}{c}{Class} \\ \cmidrule(r){3-7} 
        & & total & training & total & seen & unseen \\
        \midrule
        closed-set~\cite{tian2018audio} & AVE~\cite{tian2018audio} & 4,143 & 3,309  & 28 & 28 & 0 \\
        open-set~\cite{yu2024openave} & AVE~\cite{tian2018audio} & 4,143 & 2,505 & 28  & 21 & 7 \\
        open-vocabulary & OV-AVEBench & 24,800 & 13,182 & 67 & 46 & 21\\
        \bottomrule\hline
    \end{tabular}
    }
    \vspace{-2ex}
    \label{tab:dataset_comparison}
\end{table}

With the OV-AVEBench dataset and evaluation metrics established, we also propose \textbf{preliminary baselines} to address the OV-AVEL problem.
To facilitate the recognition of various event classes, particularly those pertaining to unseen test data, we consider leveraging the zero-shot capability of recent language-based multimodal contrastive models.
The language words are easily extendable and are not confined to predefined concepts (or categories for event classification).
By applying contrastive learning to large-scale multimodal data pairs, the resulting embeddings can capture discriminative and accurate semantics.
We opt to utilize ImageBind~\cite{girdhar2023imagebind} because it establishes a joint embedding space across multiple modalities, aligning well with the studied OV-AVEL task. 
After extracting the segment-level audio, visual, and text embeddings using ImageBind (where the text represents all potential seen and unseen event classes), we initially explore a simple \textit{\textbf{training-free baseline}}.
Specifically, we compute the cosine similarity matrices for audio-text and visual-text features, respectively.
In this way, we can identify the predicted event category for each audio and visual segment, corresponding to the highest similarity value, and subsequently generate audio-visual event predictions by verifying the consistency of the predicted audio and visual event categories.
Notably, this baseline is training-free, directly operating on the test data.
To utilize the annotated training data from the proposed OV-AVEBench dataset, we further explore a \textit{\textbf{fine-tuning baseline}}.
Although the unseen test data and training data possess distinct event categories, the temporal information in training data, which reflects the continuity of various audio-visual events, remains beneficial for the OV-AVEL task.
Inspired by this, we incorporate some lightweight transformer layers into the ImageBind model to learn temporal relations within audio and visual modalities.
Then, we fine-tune the model using the training data.
Notably, parameters of the vanilla ImageBind model remain frozen, with only those of the temporal layers being learnable; thus, the increase in resource or computational load is relatively limited.
Our experiments demonstrate that the fine-tuning baseline significantly outperforms the training-free version in inference on both seen and unseen test data.

In summary, our main contributions are three-fold:
\begin{itemize}
    \item We propose the Open-Vocabulary Audio-Visual Event Localization (OV-AVEL) task, aiming to localize both seen and unseen audio-visual events in test videos. To the best of our knowledge, this work is the first to advance the AVEL area toward more practical applications in open-vocabulary scenarios.
    \item To facilitate this new task, we construct the OV-AVEBench dataset, which includes segment-level manual event annotations. Besides, we establish standard evaluation metrics that encompass typical {accuracy}, as well as segment-level and event-level F1-scores.
    \item We present two simple baselines: one adopting a \textit{training-free} paradigm, which can be upgraded through further \textit{fine-tuning} on available training data. We hope that our benchmark will inspire future research in this field.    
\end{itemize}

\begin{figure*}[t]
  \centering
\includegraphics[width=\textwidth]{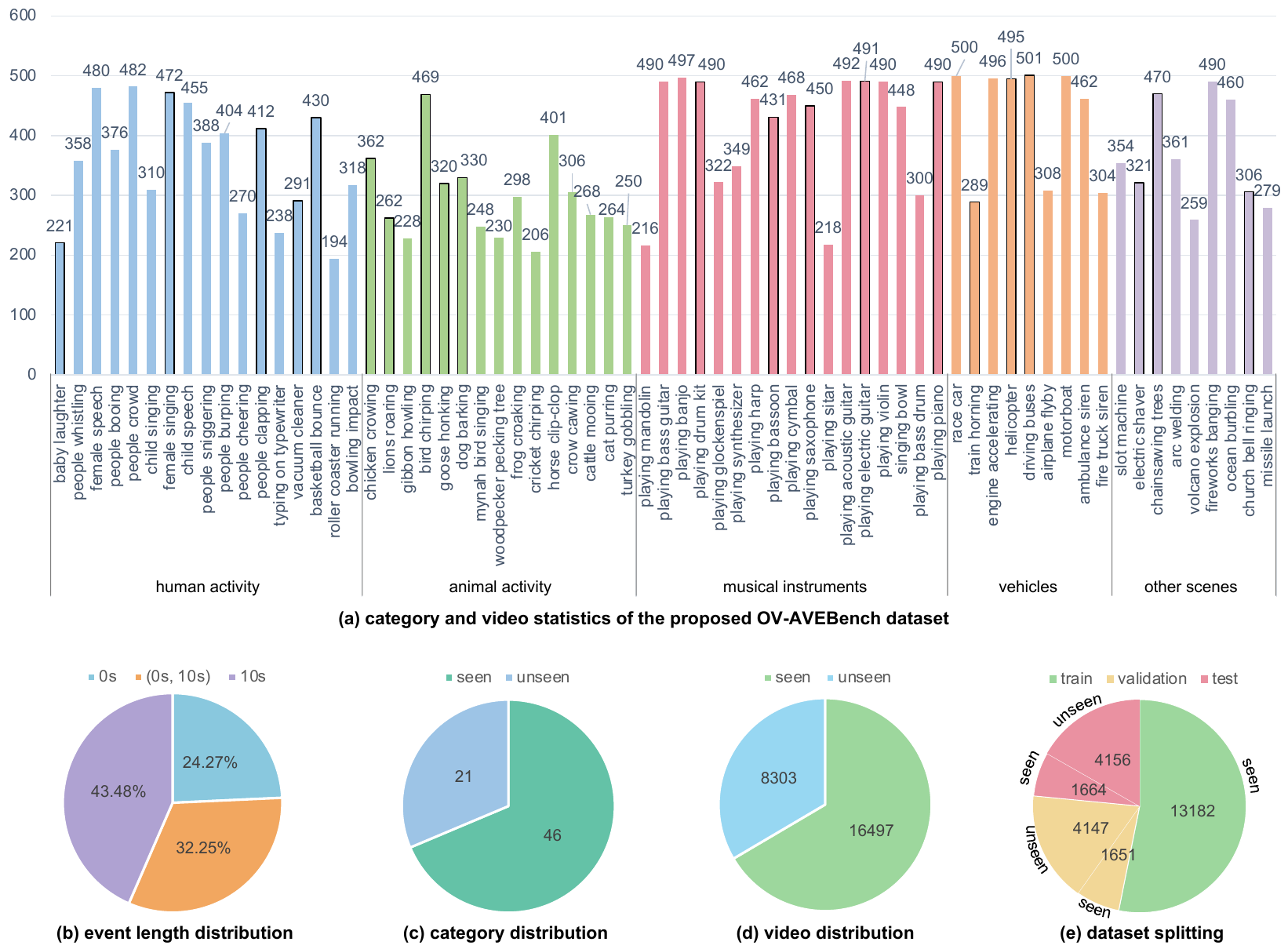}
\vspace{-4.5ex}
   \caption{\textbf{Statistics about the proposed OV-AVEBench dataset.} (a) Our OV-AVEBench contains 24,800 videos covering 67 practical audio-visual scenes from the real world.
   Each event category and its corresponding video amount are listed. The category highlighted by a black bounding box indicates that data in that category is only available during the inference phase (unseen classes/data). 
   (b) The audio-visual events in the videos exhibit various temporal scales, with some containing only background.
    We also visualize the category distribution (c), the video distribution (d) of the seen and unseen data, and the video counts for the training, validation, and test sets (e).
   }
   \vspace{-2ex}
   \label{fig:dataset_statistic}
\end{figure*}

\section{OV-AVEBench}\label{sec:ov-avebench}

We construct the OV-AVEBench dataset to facilitate the study of Open-Vocabulary Audio-Visual Event Localization (OV-AVEL).
In Table~\ref{tab:dataset_comparison}, we have provided a basic overview of OV-AVEBench.
In the following subsections, we share more details about data collection, annotation, and splitting.

\subsection{Data Collection}
The target audio-visual events in the OV-AVEL task necessitate semantic correlation between the audio and visual modalities (in at least some video segments).
To meet this requirement and avoid unnecessary costs, we resort to existing VGGSound~\cite{chen2020vggsound} dataset, a large-scale audio-visual dataset in our community that provides ample video resources.
Specifically, VGGSound dataset consists of over 200k videos covering 309 audio classes.
However, some classes may be easily recorded in audio signal but are difficult to represent in visual frames, such as \textit{wind noise} and \textit{thunder}.
Additionally, some classes are either too similar or too fine-grained (\eg, \textit{car engine starting} vs. \textit{car engine idling}), or too rare in current real-life (\eg, \textit{dinosaurs bellowing}).
We filtered out categories like these and ultimately selected {67} common and suitable classes for constructing the OV-AVEBench dataset.
The complete category list of selected categories is presented in Fig.~\ref{fig:dataset_statistic}(a).
These event categories correspond to five major groups: human activity, animal activity, musical instruments, vehicles, and several other audio-visual scenes from real life.

After determining the event categories, we downloaded the corresponding videos based on the YouTube URLs provided by the VGGSound dataset.
A small number of videos were not currently available.
Next, five volunteers were invited to manually review and check the downloadeded videos.
Some low-quality videos (\eg, those that were completely mismatched with their category tags) were further removed.
After these steps, we ultimately retained {24,800} videos as the data resources for our OV-AVEBench dataset.
The specific number of videos in each event category is shown in Fig.~\ref{fig:dataset_statistic}(a).
Each video lasts for 10 seconds.
We examine the temporal length of audio-visual events contained in the videos.
As shown in Fig.~\ref{fig:dataset_statistic}(b), we find that 43.48\% video data contain audio-visual events in all temporal segments (10 seconds), while 24.27\% contain no audio-visual events (0 seconds), \ie, containing \textit{background} class, and over one-third of the data fall in between.
These video data require the model to recognize various event categories, distinguish background segments, and localize events across different temporal scales, making our OV-AVEBench dataset applicable to the OV-AVEL task.

\subsection{Data Annotation}
After obtaining the videos, we attempt to provide audio-visual event labels for them.
For each video, we divide it into ten 1-second segments.
The intermediate video frame of each segment is extracted to represent its visual component.
The audio component is the corresponding 1-second audio sequence.
Then, the audio-visual event labels can be determined by evaluating whether the visual frame and the audio sequence describe the same event.
If they match, this segment is labeled as `1' with a meaningful event category, \eg, \textit{baby laughter}; otherwise, it is labeled as `0' and categorized as \textit{background}.
This process allows us to obtain segment-level labels.

To ensure high-quality labels for the community, we conducted the annotation process through crowdsourcing.
Ten human annotators were involved in this process.
First, we provided video examples along with guidelines to ensure that the annotators understood the annotation procedures and standards. 
Next, the formal annotation began.
After finishing the annotation, we 
exchanged annotators to perform a second-round re-evaluation.
Annotations with differing opinions were discussed to reach a final judgment.
It took us about three weeks to complete the annotation process.

\begin{figure*}[t]
  \centering
\includegraphics[width=0.85\textwidth]{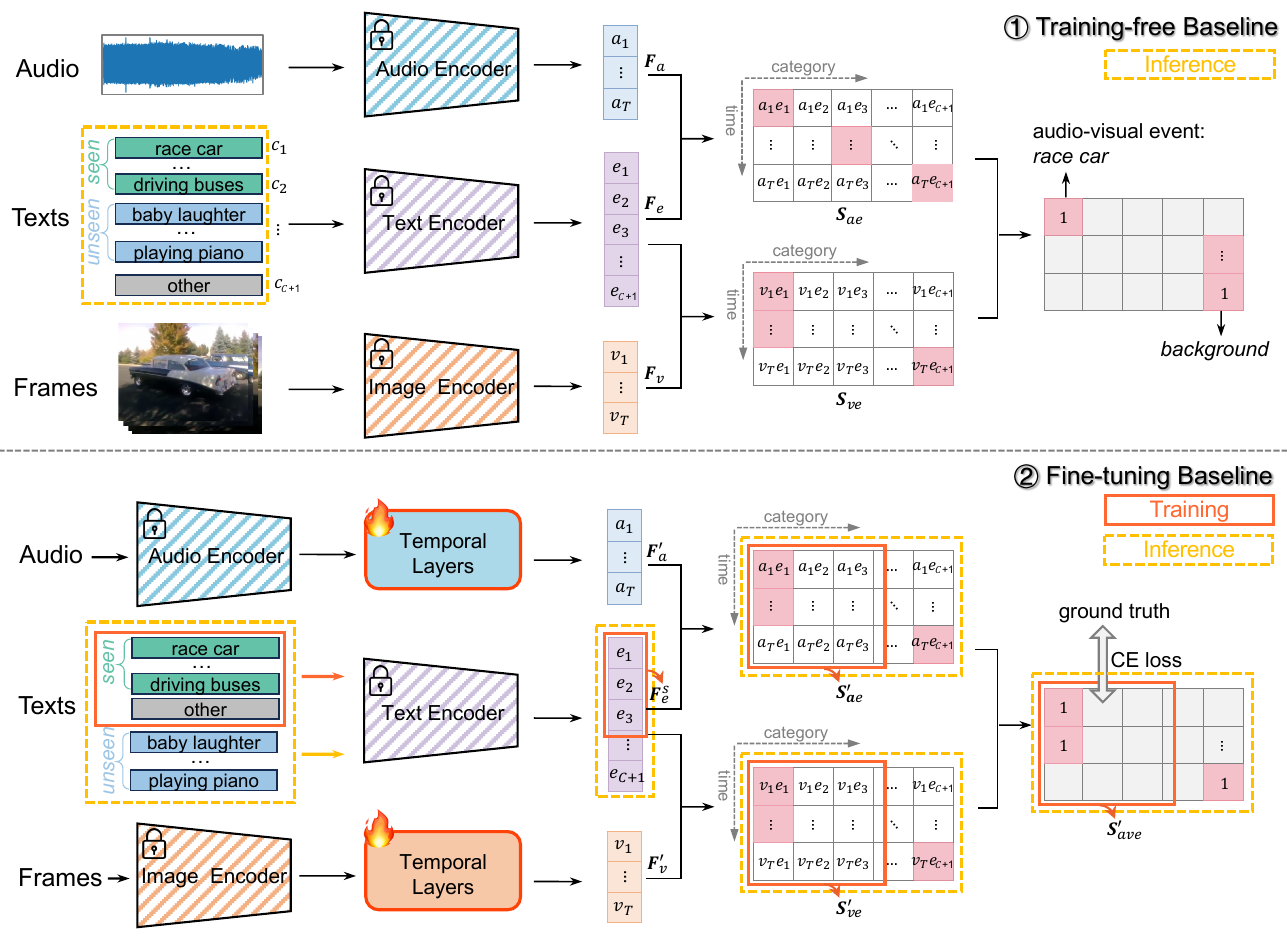}
\vspace{-3ex}
   \caption{\textbf{Overview of the proposed baseline methods.} We utilize the audio and image encoders of the pretrained Imagebind~\cite{girdhar2023imagebind} (with frozen parameters) to extract segment-level audio and visual features. 
   \ding{172} The \textit{training-free baseline} sends texts of all candidate classes (both seen and unseen) to extract features.
   Then, the audio-visual event prediction is decided by evaluating the consistency between audio-text and visual-text feature similarities.
   \ding{173} The \textit{fine-tuning baseline} additionally inserts some temporal layers into the audio and visual encoders to strengthen temporal interaction learning. 
   This model is fine-tuned/trained with training data (with seen classes).
   Only the texts of seen classes are known and used in model fine-tuning, while the unseen classes are additionally introduced during inference. 
   The final audio-visual event prediction is obtained by fusing event probabilities of audio and visual modalities.
   }
   \vspace{-2ex}
   \label{fig:framework}
\end{figure*}

\subsection{Data Splitting}
The OV-AVEL task requires handling both seen and unseen data, which means that only some event categories are present during training.
Our OV-AVEBench dataset contains a total of 67 event classes.
As shown in Fig.~\ref{fig:dataset_statistic}(c), we select 46 classes as seen classes (appearing in training data) while the remaining 21 classes as unseen (only appearing in the evaluation phase).
The detailed category names can also be observed from Fig.~\ref{fig:dataset_statistic}(a), where the unseen classes are highlighted by black bounding boxes.
Importantly, we did not simply select 21 classes from the entire category list.
Instead, we identify specific classes from each major audiovisual category group while carefully balancing the number of videos in the resulting seen and unseen data.
As shown in Fig.~\ref{fig:dataset_statistic}(d), 16,497 videos are finally used as seen data (\textit{whose classes are seen during training}), and 8,303 videos are used as unseen data. 
Notably, some of the seen data can also appear in the validation and test sets.
The detailed numbers are presented in Fig.~\ref{fig:dataset_statistic}(e).
Specifically, we split the videos into training, validation, and test sets, with respective video counts of 13,182, 5,798, and 5,820.
For the validation and test sets, we set the ratio of seen and unseen videos (\ie, \textit{videos with seen and unseen classes}) at approximately 3:7.
This allows us to evaluate models using more unseen data during the inference phase.
We will release the OV-AVEBench dataset to the community for transparency.


\section{Baselines}\label{sec:method}

\textbf{Task Formulation.} 
Given an audible video, it is divided into $T=10$ consecutive and non-overlapping segments, with $\{A_t\}_{t=1}^T$ and $\{V_t\}_{t=1}^T$ representing the audio and visual components, respectively.
The OV-AVEL task aims to localize video segments that contain an audio-visual event and identify their categories.
Each video is typically dominated by one event category.
The ground truth labels can be denoted as $\bm{Y}=\{Y_t\}_{t=1}^T \in \mathbb{R}^{T \times (C+1)}$, where `$C+1$' indicates the total number of audio-visual event classes plus a \textit{background} class.
Notably, during the inference phase, the OV-AVEL task addresses data with both seen and unseen classes.
We denote the total number of seen and unseen event classes as $C_s$ and $C_u$, respectively, where $C_s = 46$ and $C_u = 21$ in the proposed OV-AVEBench dataset. And, $C = C_s + C_u$.

\subsection{A Training-free Baseline}
The OV-AVEL task can be addressed with a straightforward training-free baseline, as illustrated in the upper part of Fig.~\ref{fig:framework}.
First, we utilize the pretrained ImageBind~\cite{girdhar2023imagebind} model discussed in Sec.~\ref{sec:intro} to extract audio and visual features.
Specifically, the sampled video frame from each visual segment is sent to the image encoder of ImageBind, yielding the segment-level visual features $\bm{F}_v = \{ \bm{v}_t \}_{t=1}^T \in \mathbb{R}^{T \times d}$, where $d=1024$ is the feature dimension.
Similarly, each audio segment is sent to the audio encoder to extract audio features, denoted as $\bm{F}_a = \{ \bm{a}_t \}_{t=1}^T \in \mathbb{R}^{T \times d}$.

Traditional approaches to closed-set AVEL~\cite{tian2018audio} typically use only the audio and visual features as model inputs for event prediction.
To achieve open-vocabulary AVEL, we adopt a zero-shot classification paradigm similar to CLIP~\cite{radford2021CLIP}.
We send all candidate event classes (seen and unseen) to the text encoder of ImageBind to obtain the text (event category) features $\bm{F}_e = \{ \bm{e}_c \}_{c=1}^{C+1} \in \mathbb{R}^{(C+1) \times d}$.
Notably, we add a special text \texttt{other} that corresponds to the \textit{background} class, handling situations that do not belong to the listed seen and unseen classes (other new potential event classes can also be flexibly added in practical applications).
Next, we compute the cosine similarities of audio-text and visual-text features, denoted as $\bm{S}_{ae} \in \mathbb{R}^{T \times (C+1)}$ and $\bm{S}_{ve} \in \mathbb{R}^{T \times (C+1)}$, as follows:
\begin{equation}
    \bm{S}_{ae} = \| \bm{F}_a \| \otimes \|  \bm{F}_e \|^{\top}, ~~
    \bm{S}_{ve} = \| \bm{F}_v \| \otimes \|  \bm{F}_e \|^{\top},
\label{eq:cosine_simm}
\end{equation}
where $\| \cdot \|$ denotes L2-normalization and $\otimes$ is the matrix multiplication. 
By scanning each row of $\bm{S}_{ae}$ and $\bm{S}_{ve}$, we can predict the category of each audio and visual segment by identifying the category with the highest cosine similarity value (marked by pink boxes shown in Fig.~\ref{fig:framework}\textbf{\ding{172}}).
The audio-visual events in target segments require that the category of the audio segment and the synchronized visual segment should be identical.
Therefore, we can easily determine the final audio-visual event predictions by checking the audio and visual class consistency for each segment:
if both modalities share the same event category, that segment contains an audio-visual event of that category; otherwise, it is recognized as \textit{background}.

\subsection{A Fine-tuning Baseline}
The baseline method described above is training-free since the parameters of the audio, image, and text encoders are frozen.
Since segment-level labels are available for the training data (with seen classes), we attempt to enhance the training-free baseline through additional fine-tuning.

ImageBind can provide advanced audiovisual features; however, they are independent at the segment level.
The temporal relations across segments are also crucial for our OV-AVEL task because the target audio-visual events typically span various temporal scales.
Motivated by this, we insert some learnable temporal layers after the audio encoder and image encoder of ImageBind to enhance the temporal interaction of each modality (illustrated in the lower part of Fig.~\ref{fig:framework}).
In practice, the temporal layers are implemented as a stack of $L$ standard Transformer~\cite{vaswani2017attention} blocks.
We denote the generated audio and visual features as $\bm{F}_a^{'} \in \mathbb{R}^{T \times d}$ and $\bm{F}_v^{'}  \in \mathbb{R}^{T \times d}$, respectively.

\textbf{Training/Fine-tuning.}
Notably, data in the training set contains only the seen classes.
Therefore, during model training/fine-tuning, only the texts of $C_s$ seen classes and additional text \texttt{other} (outlined by the orange box in Fig.~\ref{fig:framework}\textbf{\ding{173}}) are sent to the text encoder of ImageBind to extract text features, yielding $\bm{F}_{e}^{s}  \in \mathbb{R}^{ (C_s+1) \times d}$.
Then, we can compute the feature similarity matrices of audio-text and visual-text pairs, similar to Eq.~\ref{eq:cosine_simm}, denoted as $\bm{S}_{ae}^{'} \in \mathbb{R}^{T \times (C_s+1)}$ and $\bm{S}_{ve}^{'} \in \mathbb{R}^{T \times (C_s+1)}$, respectively.

The matrices $\bm{S}_{ae}^{'}$ and $\bm{S}_{ve}^{'}$ reflect the category probabilities of audio events and visual events, respectively.
We generate the final audio-visual event probability $\bm{S}_{ave}^{'} \in \mathbb{R}^{T \times (C_s+1)}$ by fusing them as follows:
\begin{equation}
    \bm{S}_{ave}^{'} = \sqrt{ \bm{S}_{ae}^{'} \odot \bm{S}_{ve}^{'} },
\label{eq:fine_tune}
\end{equation}
where $\odot$ is the Hadamard product.
This strategy differs from the direct comparison of $\bm{S}_{ae}$ and $\bm{S}_{ve}$ used in the training-free baseline, which is non-differentiable for model back-propagation.
The ground truth $\bm{Y}^{'} \in \mathbb{R}^{T \times (C_s+1)}$ for the training data can be easily obtained by selecting columns of corresponding seen classes from $\bm{Y} \in \mathbb{R}^{T \times (C+1)}$.
Then, our fine-tuning baseline is trained by optimizing the cross entropy loss between $\bm{S}_{ave}^{'}$ and $\bm{Y}^{'}$.

\textbf{Inference.} 
The OV-AVEL task involves handling both seen and unseen data (\ie, data with seen and unseen classes) during the inference phase.
As highlighted by the yellow dotted box in Fig.~\ref{fig:framework}\textbf{\ding{173}}, the texts of both seen and unseen classes are sent to the text encoder for feature extraction.
The processing of audio and visual modalities follows the same flow as in training, whereas the audio and visual segments are processed by the pretrained encoders and fine-tuned temporal layers to extract audio and visual features.
Then, we can generate the probability of audio-visual events by utilizing audio-text and visual-text feature similarities as described in Eq.~\ref{eq:fine_tune}.
The final prediction can be made by selecting the event category with the largest probability.

\section{Experiments}

\subsection{Implementation Details}
We conduct experiments on the proposed OV-AVEBench dataset and evaluate the performance of our baselines using the three evaluation metrics introduced in Sec.~\ref{sec:intro}, \ie, the accuracy (Acc.), segment-level F1-score (Seg.), and event-level F1-score (Eve.).
The average result of three metrics (\textit{Avg.}) is also reported.
In both baselines, we employ the parameters of the pretrained \texttt{ImageBind\_Huge}\footnote{\url{https://github.com/facebookresearch/ImageBind/tree/main}}, the only officially released version of the ImageBind~\cite{girdhar2023imagebind} model, to extract audio-visual-text features.
For the fine-tuning baseline, we set the batch size to 32 and fine-tune the model (learnable temporal layers) for 5 epochs; the Adam optimizer is used with a learning rate of 5e-5.
All experiments are conducted on a single NVIDIA RTX 4090D (24GB) GPU.
The source code will be released.

\begin{table*}[t]
\caption{\textbf{Benchmark on the OV-AVEBench dataset}. We report performances of our training-free and fine-tuning baselines. Additionally, we implement and compare various training-free and fine-tuning approaches.}
\vspace{-1.em}
\centering
\small
\tabcolsep 5.5pt
\begin{tabular}{c|cccc|cccc|cccc}
\hline
\multirow{2}{*}{Method} & \multicolumn{4}{c|}{seen} & \multicolumn{4}{c|}{unseen} & \multicolumn{4}{c}{total} \\ 
    & Acc. & Seg. & Eve. & \textbf{\textit{Avg.}} & Acc. & Seg. & Eve. & \textbf{\textit{Avg.}} & Acc. & Seg. & Eve. & \textbf{\textit{Avg.}} \\
    \midrule
    Video-LLaMA2~\cite{cheng2024videollama} & 50.1&40.6&32.0&40.9& 48.5&38.5&29.0&38.6& 48.9&39.1&29.8&39.3 \\
    CLIP~\cite{radford2021CLIP}\&CLAP~\cite{wu2023clap} & 51.4&41.4&31.9&41.6 & 51.6&42.2&31.6&41.8& 51.5&41.9&31.7&41.7 \\
    \textbf{Training-free (our)} & 57.5&45.0&34.0&\textbf{45.5} & 59.8&47.3&34.0&\textbf{47.0}& \textbf{59.2}&\textbf{46.7}&\textbf{34.0}&\textbf{46.6 }\\ \midrule
    {CMRA}~\cite{xu2020cross} & 65.2 & 58.8 & 54.3 & 59.4 & 36.0 & 31.0 & 26.3 & 31.1 & 44.3 & 38.9 & 34.3 & 39.2 \\
    {AVE}~\cite{tian2018audio} & 76.6 & 63.6 & 56.0 & 65.4 & 44.6 & 33.2 & 24.0 & 34.0 & 53.8 & 41.9 & 33.2 & 42.9 \\
    PSP~\cite{zhou2021positive} & 75.4 & 66.8 & 61.0 & 67.7 & 33.7 & 28.2 & 24.2 & 28.7 & 45.6 & 39.3 & 34.7 & 39.9 \\
    MM-Pyramid~\cite{yu2021mm} & {76.5} & {66.9} & {62.3} & \textbf{68.6} & 36.8 & 29.0 & 23.8 & 29.9 & 48.4 & 40.2 & 35.2 & 41.2 \\
    \textbf{Fine-tuning (our)} & 72.5 & 61.8 & 54.5 & 62.9 & {64.9} & {55.0} & {47.5} & \textbf{55.8} & \textbf{67.1} & \textbf{56.9} &\textbf{ 49.5} & \textbf{57.8} \\
        \hline
\end{tabular}
\vspace{-1.em}
\label{tab:main_results}
\end{table*}

\subsection{Main Results}
We propose both a training-free (zero-shot) baseline and a fine-tuning baseline to address the new OV-AVEL task.
Additionally, we implement several zero-shot and fine-tuning approaches to establish a comprehensive benchmark.

\noindent\textbf{Comparison of Training-free approaches.}
We compare our training-free baseline with two methods: 1) \textit{{Video-LLaMA2}}~\cite{cheng2024videollama}. We design task-specific prompts (see our \textit{supp.} for details) to enable this advanced audio-visual LLM to analyze audiovisual inputs and generate event-prediction texts.
2) \textit{{CLIP\&CLAP}}.
Instead of using ImageBind~\cite{girdhar2023imagebind}, we employ separate CLIP~\cite{radford2021CLIP} and CLAP~\cite{wu2023clap} models to extract audio-text and visual-text features, obtaining predictions following Eq.~\ref{eq:cosine_simm}.
As shown in the upper part of Table~\ref{tab:main_results}, our training-free baseline outperforms both methods.
While Video-LLaMA2 can describe events within an {entire} audio sequence, it faces challenges in achieving segment-level perception of audio-visual alignment.
Additionally, the comparison with {{CLIP\&CLAP}} highlights the advantages of using a joint feature space for multimodal feature embedding, which better captures semantic alignment among multiple modalities for the OV-AVEL task.

\noindent\textbf{Comparison of Fine-tuning approaches.} We replace the temporal layers in our fine-tuning baseline with core audio-visual fusion modules from prior closed-set AVEL methods (\textit{e.g.}, CMRA~\cite{xu2020cross}, AVE~\cite{tian2018audio}, PSP~\cite{zhou2021positive}, MM-Pyr~\cite{yu2021mm}) to \textit{enable} them perform both seen and unseen event localization for comparison.
As shown in the lower part of Table~\ref{tab:main_results}, we find that: 
1) While complex audio-visual interactions from prior methods may improve seen-class performance, they significantly degrade unseen-class performance, causing their overall results on the total test set to lag far behind ours.
2) Comparing AVE with PSP/MM-Pyr, more advanced interaction modules may exacerbate the imbalance between seen and unseen class recognition, highlighting the challenges of the OV-AVEL task.

\noindent\textbf{Training-free vs. Fine-tuning baselines.}
Our fine-tuning baseline significantly outperforms the training-free version, showing an 11.2\% improvement in the average metric (`Avg.') on the total test set.
Moreover, we observe that: the training-free baseline model performs slightly better on the unseen test data; after fine-tuning on training data, the baseline model is improved in recognizing both seen and unseen test data (17.4\%$\uparrow$ and 8.8\%$\uparrow$ in \textit{Avg.}, respectively).
The improvement is more pronounced for seen test data because their event classes have been seen in training.
However, fine-tuning remains beneficial for unseen test data.
We speculate that further fine-tuning helps the model learn temporal relations from training data, facilitating the adaption and updating of prior knowledge from ImageBind to downstream OV-AVEL task. This enables more precise localization of temporal boundaries for unseen test data.
This is supported by the event-level metric, which significantly improves from 34.0\% to 47.5\%.
We provide additional evidence and discussions on this in Sec.~\ref{sec:ablation}.
In short, the comparison between the two baselines highlights the benefits of further fine-tuning, especially when some training data with annotations are available.
More quantitative and qualitative comparison results are provided in our \textit{supp. material}.

\subsection{Ablation Studies}\label{sec:ablation}
In this section, we provide some ablation studies on the key configurations adopted in our fine-tuning baseline. More ablations are presented in our supplementary material.

\noindent\textbf{The special class text} \texttt{other}\textbf{.}
We utilize a special text \texttt{other} to assist the model in classifying events that do not belong to either the seen or unseen classes.
We conduct an ablation study to explore its impact.
As shown in Table~\ref{tab:ablation_on_other_class}, the model using additional \texttt{other} class outperforms that baseline trained without \texttt{other} by 10.8\% in average performance.
The improvement is consistent across both seen and unseen test data.
This underscores the superiority of introducing the additional class text \texttt{other}, which helps prevent the model from misclassifying unknown events or backgrounds as existing seen or unseen classes. 
In our \textit{supp. material}, we further show that the employment of \texttt{other} is slightly better than other choices like \texttt{background}.

\begin{table}[t]\scriptsize
\tabcolsep 3pt
    \centering
        \caption{\textbf{Ablation study on the employment of the text} \texttt{other}.}
    \vspace{-3ex}
    \resizebox{\linewidth}{!}{
    \begin{tabular}{l|ccc|c|ccc|c}
        \toprule
        \multirow{2}{*}{\makecell[c]{Data\\type}} & \multicolumn{4}{c|}{w. \texttt{other} }& \multicolumn{4}{c}{w/o \texttt{other} }  \\ \cmidrule(r){2-9} 
        & Acc. & Seg. & Eve. & \textit{Avg.} & Acc. & Seg. & Eve. & \textit{Avg.} \\
        \hline
        total & 67.1 & 56.9 & 49.5 & \textbf{57.8} &  59.3 & 46.9 & 34.9 & 47.0  \\
        \hline
        seen &  72.5 & 61.8 & 54.5 & \textbf{62.9} & 62.0 & 49.2 & 37.8 & 49.7 \\
        unseen & 64.9 & 55.0 & 47.5 & \textbf{55.8} & 58.2 & 46.0 & 33.7 & 45.9 \\
    \bottomrule
    \end{tabular}
    }
    \label{tab:ablation_on_other_class}
    \vspace{-3ex}
\end{table}

\begin{table}[t]\small
\tabcolsep 12pt
    \centering
    \vspace{-1ex}
        \caption{\textbf{Ablation study on the strategies for predicting $\bm{S}_{ave}^{'}$}. Detailed implementation of each strategy is shown in the main text. Results are reported on the total test data.}
    \vspace{-2ex}
     \begin{tabular}{l|ccc|c}
        \toprule
        Strategy & Acc. & Seg. & Eve. & \textit{Avg.} \\ \hline
         Prob-avg& 45.1 & 38.7 & 33.3 & 39.0 \\
         Fea-avg & 46.8 & 39.8 & 34.0 & 40.2 \\
         Sqrt (Eq.~\ref{eq:fine_tune}) & \textbf{67.1} & \textbf{56.9} & \textbf{49.5} & \textbf{57.8} \\
    \bottomrule
    \end{tabular}
\label{tab:ablation_on_ave_generation}
\vspace{-2ex}
\end{table}

\noindent\textbf{The strategy for generating $\bm{S}_{ave}^{'}$.}
In our fine-tuning baseline, we generate the audio-visual event probability $\bm{S}_{ave}^{'}$ by computing the square root of the product of predicted audio event probability $\bm{S}_{ae}^{'}$ and visual event probability $\bm{S}_{ve}^{'}$ (Eq.~\ref{eq:fine_tune}).
We refer to this strategy as \textbf{\textit{Sqrt}}.
Furthermore, we explore two additional variants to obtain $\bm{S}_{ave}^{'}$.
(1) \textbf{\textit{Prob-avg}} uses the average result of $\bm{S}_{ae}^{'}$ and $\bm{S}_{ve}^{'}$ to generate $\bm{S}_{ave}^{'}$, \ie, $\bm{S}_{ave}^{'} = (\bm{S}_{ae}^{'} + \bm{S}_{ve}^{'})/2$.
(2) \textbf{\textit{Fea-avg}} first generates the fused feature by averaging the audio feature $\bm{F}_a^{'}$ and visual feature $\bm{F}_v^{'}$, and then computes the cosine similarity (\texttt{sim}) between the fused feature and the text feature $\bm{F}_e^s$, formulated as $\bm{S}_{ave}^{'} = \texttt{sim}( \frac{\bm{F}_a^{'} + \bm{F}_v^{'}}{2}, \bm{F}_e^s)$. 
We re-train the fine-tuning baseline model using these strategies and evaluate the model on the test set.
As shown in Table~\ref{tab:ablation_on_ave_generation}, the \textbf{\textit{Sqrt}} strategy significantly outperforms the \textit{\textbf{Prob-avg}} and \textit{\textbf{Fea-avg}} variants.
The \textit{geometric mean} used by \textbf{\textit{Sqrt}} is more effective than \textit{arithmetic mean} at preventing audiovisual event predictions from being misled by a high-probability prediction in one modality.
These results indicate the importance of design in generating final audio-visual event probabilities, a factor that future research should also consider.

\begin{table}[t]\scriptsize
\tabcolsep 3pt
    \centering
        \caption{\textbf{Comparison of using temporal layers and linear layers in fine-tuning baseline.}}
    \vspace{-3ex}
    \resizebox{\linewidth}{!}{
    \begin{tabular}{l|ccc|c|ccc|c}
        \toprule
        \multirow{2}{*}{\makecell[c]{Data\\type}} & \multicolumn{4}{c|}{Temporal layer}& \multicolumn{4}{c}{ Linear layer}  \\ \cmidrule(r){2-9} 
        & Acc. & Seg. & Eve. & \textit{Avg.} & Acc. & Seg. & Eve. & \textit{Avg.} \\
        \hline
        total & 67.1 & 56.9 & 49.5 & \textbf{57.8} &  46.6 & 38.0 & 32.4 & 39.0  \\
        \hline
        seen &  72.5 & 61.8 & 54.5 & {62.9} & 76.2 & 64.5 & 56.9 & \textbf{65.8 }\\
        unseen & 64.9 & 55.0 & 47.5 & \textbf{55.8} & 34.8 & 27.4 & 22.6 & 28.3 \\
    \bottomrule
    \end{tabular}
    }
    \label{tab:temporal_linear_layers}
    \vspace{-2ex}
\end{table}

\noindent\textbf{Temporal layer vs. Linear layer.}
Our fine-tuning baseline employs some \textit{temporal layers} utilizing the self-attention mechanism in Transformer to enhance temporal interactions across video segments.
Here, we replace these with learnable \textit{linear layers} to update audio/visual features segment-wisely (\ie, without temporal interactions).
As shown in Table~\ref{tab:temporal_linear_layers}, the average performance of the model fine-tuned using linear layers lags considerably behind that using temporal layers.
Specifically, we find that the linear layers are slightly more effective than temporal layers for event localization of seen test data but are significantly inferior for unseen test data (27.5\% $\downarrow$ in \textit{Avg.} metric).
These results suggest that 1) for seen test data with classes present during training, simple linear layers may be adequate for fine-tuning; while 2) for unseen test data, sophisticated temporal relation modeling on training data becomes essential.
Consequently, developing more versatile and robust network architectures would be an intriguing area for future research.

\noindent\textbf{Different ratios of training data used for model fine-tuning.}
As shown in Table~\ref{tab:diff_ratios_of_training_data}, we fine-tune the baseline model with various ratios of training data (sampling data for each training class accordingly).
Interestingly, we find that the model achieves similar average performance across different data ratios.
For instance, using only 25\% of the training data, the model performance can reach 57.7\% in \textit{Avg.}, close to that achieved with 100\% training data.
Additionally, we find that both 100\% and 25\% of training data better improve unseen test data recognition, while using 50\% of training data is the most effective for seen test data recognition.
These results reveal a {non-linear} link between training data size and the model's performance, showing more data is not vital for seen/unseen class localization.
However, the trade-off of using less training data is the need for more training or fine-tuning epochs.
These findings suggest that determining a more balanced training strategy to optimize both seen and unseen data recognition would be a valuable direction for future work.

\begin{table}[t]\scriptsize
\tabcolsep 5pt
    \centering
        \caption{\textbf{Impact of using different ratios of training data in fine-tuning baseline.}}
    \vspace{-3ex}
    \resizebox{\linewidth}{!}{
    \begin{tabular}{c|l|ccc|c|c}
        \toprule
        \multirow{2}{*}{Training} & \multirow{2}{*}{Testing} & \multicolumn{4}{c|}{Metrics } & \multirow{2}{*}{ \makecell[c]{Best\\epoch}}  \\ \cmidrule(r){3-6} 
        & & Acc. & Seg. & Eve. & \textit{Avg.} &  \\ \hline
        \multirow{3}{*}{100\%} & total & 67.1 & 56.9 & 49.5 & \textbf{57.8 }& \multirow{3}{*}{1}  \\
        & seen & 72.5 & 61.8 & 54.5 & 62.9  \\
        & unseen & 64.9 & 55.0 & 47.5 &\textbf{ 55.8} \\
        \hline
        \multirow{3}{*}{75\%} & total & 66.1 & 56.9 & 49.9 & 57.6 & \multirow{3}{*}{3}  \\
        & seen & 75.1 & 65.4 & 59.3 & 66.6 \\
        & unseen & 62.5 & 53.5 & 46.1 & 54.0\\
        \hline
        \multirow{3}{*}{50\%} & total & 66.7 & 57.1 & 49.7 & 57.8 & \multirow{3}{*}{5} \\
        & seen & 75.3 & 66.0 & 59.9 & \textbf{67.1} \\
        & unseen & 63.5 & 53.5 & 45.6 & 54.2 \\
        \hline
        \multirow{3}{*}{25\%} & total & 66.4 & 56.8 & 49.8 & 57.7 & \multirow{3}{*}{6} \\
        & seen & 73.0 & 62.6 & 55.7 & 63.8 \\
        & unseen & 63.8 & 54.4 & 47.5 & 55.2 \\
    \bottomrule
    \end{tabular}
    }
    \label{tab:diff_ratios_of_training_data}
    \vspace{-2ex}
\end{table}

\section{Conclusion}\label{sec:conclusion}
We propose the Open-Vocabulary Audio-Visual Event Localization (OV-AVEL) task, advancing the traditional closed-set AVEL problem into more practical open-vocabulary scenarios.
Accordingly, we meticulously construct the OV-AVEBench dataset, making efforts in data collection, annotation, and splitting.
We hope that the OV-AVEBench will serve as a standardized testbed for future research on OV-AVEL.
Furthermore, we introduce two baseline approaches, a training-free baseline and a fine-tuning baseline, to address this new task.
We present some discussions based on our experimental findings, which we expect will inspire future advancements in the OV-AVEL field.

\noindent\textbf{Acknowledgement}
This work was supported in part by the National Key R\&D Program of China (NO.2024YFB3311602), the National Natural Science Foundation of China (72188101, 62272144, 62020106007, and U20A20183), and the Major Project of Anhui Province (202203a05020011, 2408085J040).

{
    \small
    \bibliographystyle{ieeenat_fullname}
    \bibliography{main}

\begin{thebibliography}{58}
\providecommand{\natexlab}[1]{#1}
\providecommand{\url}[1]{\texttt{#1}}
\expandafter\ifx\csname urlstyle\endcsname\relax
  \providecommand{\doi}[1]{doi: #1}\else
  \providecommand{\doi}{doi: \begingroup \urlstyle{rm}\Url}\fi

\bibitem[Arandjelovic and Zisserman(2017)]{arandjelovic2017look}
Relja Arandjelovic and Andrew Zisserman.
\newblock Look, listen and learn.
\newblock In \emph{ICCV}, pages 609--617, 2017.

\bibitem[Arandjelovic and Zisserman(2018)]{arandjelovic2018objects}
Relja Arandjelovic and Andrew Zisserman.
\newblock Objects that sound.
\newblock In \emph{ECCV}, pages 435--451, 2018.

\bibitem[Bao et~al.(2023)Bao, Yang, Ng, Er, and Kot]{bao2023cross}
Peijun Bao, Wenhan Yang, Boon~Poh Ng, Meng~Hwa Er, and Alex~C Kot.
\newblock Cross-modal label contrastive learning for unsupervised audio-visual event localization.
\newblock In \emph{AAAI}, pages 215--222, 2023.

\bibitem[Chen et~al.(2020)Chen, Xie, Vedaldi, and Zisserman]{chen2020vggsound}
Honglie Chen, Weidi Xie, Andrea Vedaldi, and Andrew Zisserman.
\newblock {VGGSound}: A large-scale audio-visual dataset.
\newblock In \emph{ICASSP}, pages 721--725, 2020.

\bibitem[Chen et~al.(2021)Chen, Xie, Afouras, Nagrani, Vedaldi, and Zisserman]{chen2021localizing}
Honglie Chen, Weidi Xie, Triantafyllos Afouras, Arsha Nagrani, Andrea Vedaldi, and Andrew Zisserman.
\newblock Localizing visual sounds the hard way.
\newblock In \emph{CVPR}, pages 16867--16876, 2021.

\bibitem[Chen et~al.(2024)Chen, Guo, Liu, Wu, Li, Li, and Wang]{chen2024cm}
Yaru Chen, Ruohao Guo, Xubo Liu, Peipei Wu, Guangyao Li, Zhenbo Li, and Wenwu Wang.
\newblock Cm-pie: Cross-modal perception for interactive-enhanced audio-visual video parsing.
\newblock In \emph{ICASSP}, pages 8421--8425, 2024.

\bibitem[Cheng et~al.(2024)Cheng, Leng, Zhang, Xin, Li, Chen, Zhu, Zhang, Luo, Zhao, et~al.]{cheng2024videollama}
Zesen Cheng, Sicong Leng, Hang Zhang, Yifei Xin, Xin Li, Guanzheng Chen, Yongxin Zhu, Wenqi Zhang, Ziyang Luo, Deli Zhao, et~al.
\newblock Videollama 2: Advancing spatial-temporal modeling and audio understanding in video-llms.
\newblock \emph{arXiv preprint arXiv:2406.07476}, 2024.

\bibitem[Duan et~al.(2021)Duan, Tang, Wang, Zong, Yang, and Yan]{duan2021audio}
Bin Duan, Hao Tang, Wei Wang, Ziliang Zong, Guowei Yang, and Yan Yan.
\newblock Audio-visual event localization via recursive fusion by joint co-attention.
\newblock In \emph{WACV}, pages 4013--4022, 2021.

\bibitem[Gao et~al.(2023)Gao, Chen, and Xu]{gao2023collecting}
Junyu Gao, Mengyuan Chen, and Changsheng Xu.
\newblock Collecting cross-modal presence-absence evidence for weakly-supervised audio-visual event perception.
\newblock In \emph{CVPR}, pages 18827--18836, 2023.

\bibitem[Ge et~al.(2023)Ge, Jiang, Yin, Wang, Cheng, and Gu]{ge2023learning}
Shiping Ge, Zhiwei Jiang, Yafeng Yin, Cong Wang, Zifeng Cheng, and Qing Gu.
\newblock Learning event-specific localization preferences for audio-visual event localization.
\newblock In \emph{ACM MM}, pages 3446--3454, 2023.

\bibitem[Girdhar et~al.(2023)Girdhar, El-Nouby, Liu, Singh, Alwala, Joulin, and Misra]{girdhar2023imagebind}
Rohit Girdhar, Alaaeldin El-Nouby, Zhuang Liu, Mannat Singh, Kalyan~Vasudev Alwala, Armand Joulin, and Ishan Misra.
\newblock Imagebind: One embedding space to bind them all.
\newblock In \emph{CVPR}, pages 15180--15190, 2023.

\bibitem[Guo et~al.(2023)Guo, Ying, Chen, Niu, Li, Qu, Qi, Zhou, Xing, Yue, Shi, Wang, Zhang, and Liang]{guo2023avis}
Ruohao Guo, Xianghua Ying, Yaru Chen, Dantong Niu, Guangyao Li, Liao Qu, Yanyu Qi, Jinxing Zhou, Bowei Xing, Wenzhen Yue, Ji Shi, Qixun Wang, Peiliang Zhang, and Buwen Liang.
\newblock Audio-visual instance segmentation.
\newblock \emph{arXiv preprint arXiv:2310.18709}, 2023.

\bibitem[Guo et~al.(2024{\natexlab{a}})Guo, Niu, Qu, Qi, Shi, Yue, Xing, Chen, and Ying]{guo2024instance}
Ruohao Guo, Dantong Niu, Liao Qu, Yanyu Qi, Ji Shi, Wenzhen Yue, Bowei Xing, Taiyan Chen, and Xianghua Ying.
\newblock Instance-level panoramic audio-visual saliency detection and ranking.
\newblock In \emph{ACM MM}, pages 9426--9434, 2024{\natexlab{a}}.

\bibitem[Guo et~al.(2024{\natexlab{b}})Guo, Qu, Niu, Qi, Yue, Shi, Xing, and Ying]{guo2024open}
Ruohao Guo, Liao Qu, Dantong Niu, Yanyu Qi, Wenzhen Yue, Ji Shi, Bowei Xing, and Xianghua Ying.
\newblock Open-vocabulary audio-visual semantic segmentation.
\newblock In \emph{ACM MM}, pages 7533--7541, 2024{\natexlab{b}}.

\bibitem[Guo et~al.(2024{\natexlab{c}})Guo, Ying, Qi, and Qu]{guo2024unitr}
Ruohao Guo, Xianghua Ying, Yanyu Qi, and Liao Qu.
\newblock Unitr: A unified transformer-based framework for co-object and multi-modal saliency detection.
\newblock \emph{IEEE transactions on multimedia}, 2024{\natexlab{c}}.

\bibitem[Hao et~al.(2024)Hao, Mao, He, Han, Dai, and Zhong]{hao2024improving}
Dawei Hao, Yuxin Mao, Bowen He, Xiaodong Han, Yuchao Dai, and Yiran Zhong.
\newblock Improving audio-visual segmentation with bidirectional generation.
\newblock In \emph{AAAI}, pages 2067--2075, 2024.

\bibitem[He et~al.(2024)He, Liu, Li, Zhao, Shen, Kong, Yang, and Zeng]{he2024cace}
Xiang He, Xiangxi Liu, Yang Li, Dongcheng Zhao, Guobin Shen, Qingqun Kong, Xin Yang, and Yi Zeng.
\newblock Cace-net: Co-guidance attention and contrastive enhancement for effective audio-visual event localization.
\newblock \emph{arXiv preprint arXiv:2408.01952}, 2024.

\bibitem[Li et~al.(2024{\natexlab{a}})Li, Yang, Mao, Ye, Chen, and Zhong]{li2024tri}
Bingliang Li, Fengyu Yang, Yuxin Mao, Qingwen Ye, Hongkai Chen, and Yiran Zhong.
\newblock Tri-ergon: Fine-grained video-to-audio generation with multi-modal conditions and lufs control.
\newblock \emph{arXiv preprint arXiv:2412.20378}, 2024{\natexlab{a}}.

\bibitem[Li et~al.(2022)Li, Wei, Tian, Xu, Wen, and Hu]{li2022learning}
Guangyao Li, Yake Wei, Yapeng Tian, Chenliang Xu, Ji-Rong Wen, and Di Hu.
\newblock Learning to answer questions in dynamic audio-visual scenarios.
\newblock In \emph{CVPR}, pages 19108--19118, 2022.

\bibitem[Li et~al.(2023)Li, Hou, and Hu]{li2023progressive}
Guangyao Li, Wenxuan Hou, and Di Hu.
\newblock Progressive spatio-temporal perception for audio-visual question answering.
\newblock In \emph{ACM MM}, pages 7808--7816, 2023.

\bibitem[Li et~al.(2024{\natexlab{b}})Li, Guo, Zhou, Zhang, and Wang]{li2024object}
Zhangbin Li, Dan Guo, Jinxing Zhou, Jing Zhang, and Meng Wang.
\newblock Object-aware adaptive-positivity learning for audio-visual question answering.
\newblock In \emph{AAAI}, pages 3306--3314, 2024{\natexlab{b}}.

\bibitem[Li et~al.(2024{\natexlab{c}})Li, Zhou, Zhang, Tang, Li, and Guo]{li2024patch}
Zhangbin Li, Jinxing Zhou, Jing Zhang, Shengeng Tang, Kun Li, and Dan Guo.
\newblock Patch-level sounding object tracking for audio-visual question answering.
\newblock \emph{arXiv preprint arXiv:2412.10749}, 2024{\natexlab{c}}.

\bibitem[Lin et~al.(2019)Lin, Li, and Wang]{lin2019dual}
Yan-Bo Lin, Yu-Jhe Li, and Yu-Chiang~Frank Wang.
\newblock Dual-modality seq2seq network for audio-visual event localization.
\newblock In \emph{ICASSP}, pages 2002--2006, 2019.

\bibitem[Mahmud and Marculescu(2023)]{mahmud2023ave}
Tanvir Mahmud and Diana Marculescu.
\newblock Ave-clip: Audioclip-based multi-window temporal transformer for audio visual event localization.
\newblock In \emph{WACV}, pages 5158--5167, 2023.

\bibitem[Mahmud et~al.(2024)Mahmud, Tian, and Marculescu]{mahmud2024t}
Tanvir Mahmud, Yapeng Tian, and Diana Marculescu.
\newblock T-vsl: Text-guided visual sound source localization in mixtures.
\newblock In \emph{CVPR}, pages 26742--26751, 2024.

\bibitem[Mao et~al.(2023{\natexlab{a}})Mao, Zhang, Xiang, Lv, Zhong, and Dai]{mao2023contrastive}
Yuxin Mao, Jing Zhang, Mochu Xiang, Yunqiu Lv, Yiran Zhong, and Yuchao Dai.
\newblock Contrastive conditional latent diffusion for audio-visual segmentation.
\newblock \emph{arXiv preprint arXiv:2307.16579}, 2023{\natexlab{a}}.

\bibitem[Mao et~al.(2023{\natexlab{b}})Mao, Zhang, Xiang, Zhong, and Dai]{mao2023multimodal}
Yuxin Mao, Jing Zhang, Mochu Xiang, Yiran Zhong, and Yuchao Dai.
\newblock Multimodal variational auto-encoder based audio-visual segmentation.
\newblock In \emph{ICCV}, pages 954--965, 2023{\natexlab{b}}.

\bibitem[Mao et~al.(2024)Mao, Shen, Zhang, Qin, Zhou, Xiang, Zhong, and Dai]{mao2024tavgbench}
Yuxin Mao, Xuyang Shen, Jing Zhang, Zhen Qin, Jinxing Zhou, Mochu Xiang, Yiran Zhong, and Yuchao Dai.
\newblock Tavgbench: Benchmarking text to audible-video generation.
\newblock In \emph{ACM MM}, pages 6607--6616, 2024.

\bibitem[Mazumder et~al.(2021)Mazumder, Singh, Parida, and Namboodiri]{mazumder2021avgzslnet}
Pratik Mazumder, Pravendra Singh, Kranti~Kumar Parida, and Vinay~P Namboodiri.
\newblock Avgzslnet: Audio-visual generalized zero-shot learning by reconstructing label features from multi-modal embeddings.
\newblock In \emph{WACV}, pages 3090--3099, 2021.

\bibitem[Mercea et~al.(2022{\natexlab{a}})Mercea, Hummel, Koepke, and Akata]{mercea2022temporal}
Otniel-Bogdan Mercea, Thomas Hummel, A~Sophia Koepke, and Zeynep Akata.
\newblock Temporal and cross-modal attention for audio-visual zero-shot learning.
\newblock In \emph{ECCV}, pages 488--505. Springer, 2022{\natexlab{a}}.

\bibitem[Mercea et~al.(2022{\natexlab{b}})Mercea, Riesch, Koepke, and Akata]{mercea2022audio}
Otniel-Bogdan Mercea, Lukas Riesch, A Koepke, and Zeynep Akata.
\newblock Audio-visual generalised zero-shot learning with cross-modal attention and language.
\newblock In \emph{CVPR}, pages 10553--10563, 2022{\natexlab{b}}.

\bibitem[Mo and Morgado(2022)]{mo2022localizing}
Shentong Mo and Pedro Morgado.
\newblock Localizing visual sounds the easy way.
\newblock In \emph{ECCV}, pages 218--234. Springer, 2022.

\bibitem[Mo and Morgado(2024)]{mo2024audio}
Shentong Mo and Pedro Morgado.
\newblock Audio-visual generalized zero-shot learning the easy way.
\newblock \emph{arXiv preprint arXiv:2407.13095}, 2024.

\bibitem[Munasinghe et~al.(2023)Munasinghe, Thushara, Maaz, Rasheed, Khan, Shah, and Khan]{munasinghe2023pg}
Shehan Munasinghe, Rusiru Thushara, Muhammad Maaz, Hanoona~Abdul Rasheed, Salman Khan, Mubarak Shah, and Fahad Khan.
\newblock Pg-video-llava: Pixel grounding large video-language models.
\newblock \emph{arXiv preprint arXiv:2311.13435}, 2023.

\bibitem[Parida et~al.(2020)Parida, Matiyali, Guha, and Sharma]{parida2020coordinated}
Kranti Parida, Neeraj Matiyali, Tanaya Guha, and Gaurav Sharma.
\newblock Coordinated joint multimodal embeddings for generalized audio-visual zero-shot classification and retrieval of videos.
\newblock In \emph{WACV}, pages 3251--3260, 2020.

\bibitem[Qian et~al.(2020)Qian, Hu, Dinkel, Wu, Xu, and Lin]{qian2020multiple}
Rui Qian, Di Hu, Heinrich Dinkel, Mengyue Wu, Ning Xu, and Weiyao Lin.
\newblock Multiple sound sources localization from coarse to fine.
\newblock In \emph{ECCV}, pages 292--308, 2020.

\bibitem[Radford et~al.(2021)Radford, Kim, Hallacy, Ramesh, Goh, Agarwal, Sastry, Askell, Mishkin, Clark, et~al.]{radford2021CLIP}
Alec Radford, Jong~Wook Kim, Chris Hallacy, Aditya Ramesh, Gabriel Goh, Sandhini Agarwal, Girish Sastry, Amanda Askell, Pamela Mishkin, Jack Clark, et~al.
\newblock Learning transferable visual models from natural language supervision.
\newblock In \emph{ICML}, pages 8748--8763, 2021.

\bibitem[Shen et~al.(2023)Shen, Li, Zhou, Qin, He, Han, Li, Dai, Kong, Wang, et~al.]{shen2023fine}
Xuyang Shen, Dong Li, Jinxing Zhou, Zhen Qin, Bowen He, Xiaodong Han, Aixuan Li, Yuchao Dai, Lingpeng Kong, Meng Wang, et~al.
\newblock Fine-grained audible video description.
\newblock In \emph{CVPR}, pages 10585--10596, 2023.

\bibitem[Tian et~al.(2018)Tian, Shi, Li, Duan, and Xu]{tian2018audio}
Yapeng Tian, Jing Shi, Bochen Li, Zhiyao Duan, and Chenliang Xu.
\newblock Audio-visual event localization in unconstrained videos.
\newblock In \emph{ECCV}, pages 247--263, 2018.

\bibitem[Tian et~al.(2020)Tian, Li, and Xu]{tian2020unified}
Yapeng Tian, Dingzeyu Li, and Chenliang Xu.
\newblock Unified multisensory perception: Weakly-supervised audio-visual video parsing.
\newblock In \emph{ECCV}, pages 436--454, 2020.

\bibitem[Vaswani et~al.(2017)Vaswani, Shazeer, Parmar, Uszkoreit, Jones, Gomez, Kaiser, and Polosukhin]{vaswani2017attention}
Ashish Vaswani, Noam Shazeer, Niki Parmar, Jakob Uszkoreit, Llion Jones, Aidan~N Gomez, {\L}ukasz Kaiser, and Illia Polosukhin.
\newblock Attention is all you need.
\newblock In \emph{NeurIPS}, pages 1--11, 2017.

\bibitem[Wu et~al.(2019)Wu, Zhu, Yan, and Yang]{wu2019dual}
Yu Wu, Linchao Zhu, Yan Yan, and Yi Yang.
\newblock Dual attention matching for audio-visual event localization.
\newblock In \emph{ICCV}, pages 6292--6300, 2019.

\bibitem[Wu et~al.(2022)Wu, Zhang, Wang, and Huang]{wu2022span}
Yiling Wu, Xinfeng Zhang, Yaowei Wang, and Qingming Huang.
\newblock Span-based audio-visual localization.
\newblock In \emph{ACM MM}, pages 1252--1260, 2022.

\bibitem[Wu et~al.(2023)Wu, Chen, Zhang, Hui, Berg-Kirkpatrick, and Dubnov]{wu2023clap}
Yusong Wu, Ke Chen, Tianyu Zhang, Yuchen Hui, Taylor Berg-Kirkpatrick, and Shlomo Dubnov.
\newblock Large-scale contrastive language-audio pretraining with feature fusion and keyword-to-caption augmentation.
\newblock In \emph{ICASSP}, pages 1--5, 2023.

\bibitem[Xia and Zhao(2022)]{xia2022cross}
Yan Xia and Zhou Zhao.
\newblock Cross-modal background suppression for audio-visual event localization.
\newblock In \emph{CVPR}, pages 19989--19998, 2022.

\bibitem[Xu et~al.(2020)Xu, Zeng, Wu, Tan, and Gan]{xu2020cross}
Haoming Xu, Runhao Zeng, Qingyao Wu, Mingkui Tan, and Chuang Gan.
\newblock Cross-modal relation-aware networks for audio-visual event localization.
\newblock In \emph{ACM MM}, pages 3893--3901, 2020.

\bibitem[Yang et~al.(2022)Yang, Wang, Duan, Chen, Hou, Jin, and Zhu]{yang2022avqa}
Pinci Yang, Xin Wang, Xuguang Duan, Hong Chen, Runze Hou, Cong Jin, and Wenwu Zhu.
\newblock Avqa: A dataset for audio-visual question answering on videos.
\newblock In \emph{ACM MM}, pages 3480--3491, 2022.

\bibitem[Yu et~al.(2022)Yu, Cheng, Zhao, Feng, and Zhang]{yu2021mm}
Jiashuo Yu, Ying Cheng, Rui-Wei Zhao, Rui Feng, and Yuejie Zhang.
\newblock {MM-Pyramid}: Multimodal pyramid attentional network for audio-visual event localization and video parsing.
\newblock In \emph{ACM MM}, pages 6241--6249, 2022.

\bibitem[Yu et~al.(2024)Yu, Zhang, Teng, and Fan]{yu2024openave}
Jiale Yu, Baopeng Zhang, Zhu Teng, and Jianping Fan.
\newblock Open{AVE}: Moving towards open set audio-visual event localization.
\newblock In \emph{ACM MM}, pages 1--10, 2024.

\bibitem[Zhao et~al.(2024)Zhao, Zhou, Guo, Zhao, and Chen]{zhao2024multimodal}
Pengcheng Zhao, Jinxing Zhou, Dan Guo, Yang Zhao, and Yanxiang Chen.
\newblock Multimodal class-aware semantic enhancement network for audio-visual video parsing.
\newblock \emph{arXiv preprint arXiv:2412.11248}, 2024.

\bibitem[Zhou et~al.(2021)Zhou, Zheng, Zhong, Hao, and Wang]{zhou2021positive}
Jinxing Zhou, Liang Zheng, Yiran Zhong, Shijie Hao, and Meng Wang.
\newblock Positive sample propagation along the audio-visual event line.
\newblock In \emph{CVPR}, pages 8436--8444, 2021.

\bibitem[Zhou et~al.(2022{\natexlab{a}})Zhou, Guo, and Wang]{zhou2022cpsp}
Jinxing Zhou, Dan Guo, and Meng Wang.
\newblock Contrastive positive sample propagation along the audio-visual event line.
\newblock \emph{TPAMI}, pages 1--18, 2022{\natexlab{a}}.

\bibitem[Zhou et~al.(2022{\natexlab{b}})Zhou, Wang, Zhang, Sun, Zhang, Birchfield, Guo, Kong, Wang, and Zhong]{zhou2022avs}
Jinxing Zhou, Jianyuan Wang, Jiayi Zhang, Weixuan Sun, Jing Zhang, Stan Birchfield, Dan Guo, Lingpeng Kong, Meng Wang, and Yiran Zhong.
\newblock Audio--visual segmentation.
\newblock In \emph{ECCV}, pages 386--403, 2022{\natexlab{b}}.

\bibitem[Zhou et~al.(2023{\natexlab{a}})Zhou, Guo, Zhong, and Wang]{zhou2023improving}
Jinxing Zhou, Dan Guo, Yiran Zhong, and Meng Wang.
\newblock Improving audio-visual video parsing with pseudo visual labels.
\newblock \emph{arXiv preprint arXiv:2303.02344}, 2023{\natexlab{a}}.

\bibitem[Zhou et~al.(2023{\natexlab{b}})Zhou, Shen, Wang, Zhang, Sun, Zhang, Birchfield, Guo, Kong, Wang, et~al.]{zhou2023avss}
Jinxing Zhou, Xuyang Shen, Jianyuan Wang, Jiayi Zhang, Weixuan Sun, Jing Zhang, Stan Birchfield, Dan Guo, Lingpeng Kong, Meng Wang, et~al.
\newblock Audio-visual segmentation with semantics.
\newblock \emph{arXiv preprint arXiv:2301.13190}, 2023{\natexlab{b}}.

\bibitem[Zhou et~al.(2024{\natexlab{a}})Zhou, Guo, Mao, Zhong, Chang, and Wang]{zhou2024label}
Jinxing Zhou, Dan Guo, Yuxin Mao, Yiran Zhong, Xiaojun Chang, and Meng Wang.
\newblock Label-anticipated event disentanglement for audio-visual video parsing.
\newblock In \emph{ECCV}, pages 1--22, 2024{\natexlab{a}}.

\bibitem[Zhou et~al.(2024{\natexlab{b}})Zhou, Guo, Zhong, and Wang]{zhou2024vaplan}
Jinxing Zhou, Dan Guo, Yiran Zhong, and Meng Wang.
\newblock Advancing weakly-supervised audio-visual video parsing via segment-wise pseudo labeling.
\newblock \emph{IJCV}, pages 1--22, 2024{\natexlab{b}}.

\bibitem[Zhou et~al.(2024{\natexlab{c}})Zhou, Zhou, Qian, Tang, Chang, and Guo]{zhou2024dense}
Ziheng Zhou, Jinxing Zhou, Wei Qian, Shengeng Tang, Xiaojun Chang, and Dan Guo.
\newblock Dense audio-visual event localization under cross-modal consistency and multi-temporal granularity collaboration.
\newblock \emph{arXiv preprint arXiv:2412.12628}, 2024{\natexlab{c}}.

\end{thebibliography}
}

\newpage
\appendix
\renewcommand{\thetable}{A\arabic{table}}
\renewcommand{\thefigure}{A\arabic{figure}}

\begin{table}[t]\small
\tabcolsep 16pt
    \centering
    \vspace{-1ex}
        \caption{\textbf{Ablation study on the number of temporal layers $L$.} Results are reported on the total test data.}
    \vspace{-3ex}
     \begin{tabular}{l|ccc|c}
        \toprule
        $L$ & Acc. & Seg. & Eve. & \textit{Avg.} \\ \hline
        1 &  \textbf{67.1} & \textbf{56.9} & \textbf{49.5} & \textbf{57.8} \\ 
        2 &  65.4 & {56.0} & 49.2 & 56.9 \\
        3 &  62.8 & 54.0 & {47.3} & 54.7 \\
    \bottomrule
    \end{tabular}
\label{tab:ablation_on_temporal_layers}
\end{table}

\section{The number of temporal layers  $L$}
Our fine-tuning baseline employs $L$ learnable temporal layers to enhance temporal interactions within audio and visual modalities.
The results, as shown in Table~\ref{tab:ablation_on_temporal_layers}, illustrate the impacts of varying the number of layers.
The model achieves the highest average performance using only one temporal layer.
Increasing the number of temporal layers may make the model more complex and lead to overfitting, thus degrading the performance.
Consequently, we identify $L=1$ to implement the temporal layer, which is lightweight and only introduces 8.4M trainable parameters.

\begin{table}[t]\scriptsize
\tabcolsep 3pt
    \centering
        \caption{\textbf{Ablation study on the employment of the text} \texttt{other}. `TF' and `FT' represent the training-free baseline and fine-tuning baseline, respectively.}
    \vspace{-3ex}
    \resizebox{\linewidth}{!}{
    \begin{tabular}{l|l|ccc|c|ccc|c}
        \toprule
        \multirow{5}{*}{TF} & \multirow{2}{*}{\makecell[c]{Data\\type}} & \multicolumn{4}{c|}{w. \texttt{other} }& \multicolumn{4}{c}{w. \texttt{background} }  \\ \cmidrule(r){3-10} 
        & & Acc. & Seg. & Eve. & \textit{Avg.} & Acc. & Seg. & Eve. & \textit{Avg.} \\
        \cmidrule(r){2-10} 
        & total & 59.2 & 46.7 & 34.0 & \textbf{46.6} & 59.1 & 46.6 & 33.8 & 46.5  \\
        \cmidrule(r){2-10} 
         & seen &  57.5 & 45.0 & 34.0 & \textbf{45.5} & 57.5 & 45.1 & 34.0 & \textbf{45.5} \\
         & unseen & 59.8 & 47.3 & 34.0 & \textbf{47.0} & 59.7 & 47.2 & 33.7 & 46.9 \\
         \midrule
        \multirow{5}{*}{FT} & \multirow{2}{*}{\makecell[c]{Data\\type}} & \multicolumn{4}{c|}{w. \texttt{other} }& \multicolumn{4}{c}{w. \texttt{background} }  \\ \cmidrule(r){3-10} 
        & & Acc. & Seg. & Eve. & \textit{Avg.} & Acc. & Seg. & Eve. & \textit{Avg.} \\
        \cmidrule(r){2-10} 
        & total & 67.1 & 56.9 & 49.5 & \textbf{57.8} &  66.2 & 56.1 & 48.5 & 56.9  \\
        \cmidrule(r){2-10} 
         & seen &  72.5 & 61.8 & 54.5 & \textbf{62.9} & 71.8 & 60.9 & 53.7 & 62.1 \\
         & unseen & 64.9 & 55.0 & 47.5 & \textbf{55.8} & 63.9 & 54.1 & 46.4 & 54.8 \\
    \bottomrule
    \end{tabular}
    }
    \label{tab:ablation_on_other_class}
\end{table}

\section{Further Ablation Study on \texttt{other}}
In Sec. 4.3 of our main paper, we have demonstrated that our baseline models using additional class text \texttt{other} outperform models that do not use \texttt{other}.
Here, we further compare the employment of \texttt{other} with another option, namely \texttt{background}.
The experimental results are shown in Table~\ref{tab:ablation_on_other_class}.
For both the training-free and fine-tuning baselines, the use of \texttt{other} is slightly better than \texttt{background}.
Compared to \texttt{background}, we speculate that the text \texttt{other} can further help the model deal with situations that include other meaningful event classes not listed in the seen and unseen class texts.

\begin{table}[t]\small
\tabcolsep 12pt
    \centering
    \vspace{-1ex}
        \caption{\textbf{Temporal interactions in intra- and cross- modalities for model fine-tuning.} Results are reported on the total test data.}
    \vspace{-2ex}
    \begin{tabular}{l|ccc|c}
        \toprule
        Cases & Acc. & Seg. & Eve. & \textit{Avg.} \\ \hline
        \textbf{intra only} &  \textbf{67.1} & \textbf{56.9} & \textbf{49.5} & \textbf{57.8} \\ 
        cross only &  54.4 & 45.9 & 39.2 & 46.5 \\
        intra + cross & 63.5 & 54.3 & 47.1 & 55.0 \\
    \bottomrule
    \end{tabular}
    \label{tab:ablation_on_intra_cross_attention}
\end{table}

\section{Intra-modal vs. Cross-modal temporal layers}
The temporal layers in our fine-tuning baseline facilitate temporal interactions within the audio and visual modalities (\textit{intra-modal}).
We also attempted to insert some temporal layers to capture \textit{cross-modal} temporal relations.
As shown in Table~\ref{tab:ablation_on_intra_cross_attention}, adding cross-modal temporal layers does not yield improvements.
We speculate that the audio and visual features extracted by the pretrained ImageBind model can provide explicit and precise semantics of audio events and visual events, reducing the need for cross-modal interactions.
By focusing on the temporal interactions in intra-modality, the model can achieve satisfactory performance.

\begin{table}[t]\scriptsize
\tabcolsep 3pt
    \centering
        \caption{\textbf{Comparison between the Training-free baseline with the variant CLIP\&CLAP.} 
        The default implementation in our main paper uses ImageBind~\cite{girdhar2023imagebind} to \textit{jointly} extract multimodal features and generate audio-visual event predictions. In contrast, the \textit{separate} variant uses the pretrained CLAP~\cite{wu2023clap} and CLIP~\cite{radford2021CLIP} models to extract features independently and computes the audio-text and visual-text feature similarities separately.}
        \vspace{-2ex}
    \resizebox{\linewidth}{!}{
    \begin{tabular}{l|ccc|c|ccc|c}
        \toprule
        \multirow{2}{*}{\makecell[c]{Data\\type}} & \multicolumn{4}{c|}{ImageBind (\textit{joint})}& \multicolumn{4}{c}{CLAP\&CLIP (\textit{separate})}  \\ \cmidrule(r){2-9} 
        & Acc. & Seg. & Eve. & \textit{Avg.} & Acc. & Seg. & Eve. & \textit{Avg.} \\
        \hline
        total & 59.2 & 46.7 & 34.0 & \textbf{46.6} & 51.5 & 41.9 & 31.7 & 41.7 \\
        \hline
        seen & 57.5 & 45.0 & 34.0 &\textbf{45.5} &  51.4 & 41.4 & 31.9 & 41.6 \\
        unseen & 59.8 & 47.3 & 34.0 &\textbf{47.0} & 51.6 & 42.2 & 31.6 & 41.8 \\
    \bottomrule
    \end{tabular}
    }
    \label{tab:joint_clipclap}
\end{table}

\section{Comparison with the CLIP\&CLAP}
In sec. 4.2, we compare our training-free baseline with another zero-shot approach, CLIP\&CLAP. Here, we provide more implementation details.
The training-free baseline introduced in our main paper utilizes ImageBind~\cite{girdhar2023imagebind} to extract audio, visual, and textual embeddings. It computes the audio-text and visual-text feature (cosine) similarities to determine final audio-visual event predictions.
We refer to this strategy as \textbf{\textit{joint}} since multimodal features are extracted from a shared feature space.
Furthermore, we compare this approach with another variant, where the audio-text and visual-text feature similarities are calculated using feature embeddings from \textbf{\textit{separate}} backbones.
Specifically, for each segment, the pretrained CLAP~\cite{wu2023clap} model is used to extract the audio and text features to generate the audio-text feature similarity; the pretrained CLIP~\cite{radford2021CLIP} model is used to extract the visual and text features to generate the visual-text feature similarity.
Notably, the text encoders of CLAP and CLIP models are different, so the text features are extracted independently.
After obtaining the audio-text and visual-text feature similarities, we identify the event categories of the audio segments and visual segments based on the highest similarity values.
The final audio-visual event prediction can be made by comparing the consistency of the predicted audio and visual event categories.
The experimental results are shown in Table~\ref{tab:joint_clipclap}.
The \textbf{\textit{joint}} baseline model using ImageBind significantly outperforms the \textit{\textbf{separate}} variant, with improvements of 4.9\%, 3.9\%, and 5.2\% in the \textit{Avg.} metric on the total, seen, and unseen test data, respectively.
These results indicate the advantages of adopting a joint feature space for multimodal feature embedding, which can better capture semantic alignment among multiple modalities for the OV-AVEL task.

\section{Zero-shot Evaluation on AVE~\cite{tian2018audio} Dataset}
The AVE dataset is constructed for the closed-set AVEL task~\cite{tian2018audio}.
Here, we directly apply our two baseline models to the test set of AVE dataset in a zero-shot inference manner.
As shown in Table~\ref{tab:exp_on_ave_dataset}, the fine-tuning baseline continues to outperform the training-free version, demonstrating results competitive with the prior unsupervised state-of-the-art (SOTA) method CMLCL~\cite{bao2023cross}.
Notably, CMLCL still uses unlabeled videos of the training set in the AVE dataset.
Moreover, if further fine-tuning our baseline model on the AVE dataset, the model can reach 79.6\% accuracy without sophisticated designs, approaching the performance of fully-supervised AVEL methods~\cite{tian2018audio,yu2021mm,zhou2022cpsp,xia2022cross,ge2023learning}.
Nevertheless, we encourage readers to focus on the intrinsic differences: our method is designed for the open-vocabulary AVEL, while prior SOTA methods are tailored specifically for closed-set AVEL.

\begin{table}[t]\small
\tabcolsep 12pt
    \centering
        \caption{\textbf{Zero-shot evaluation on AVE~\cite{tian2018audio} dataset.}}
    \vspace{-2ex}
    \begin{tabular}{l|c|c}
        \toprule
        Manners & Methods & Acc. \\ \hline
        \multirow{2}{*}{zero-shot} &  training-free (our) &  54.8 \\ 
        & fine-tuning (our) & 61.9 \\ \hline
        {\color{gray}{unsupervised}} & {\color{gray}CMLCL~\cite{bao2023cross}} & {\color{gray}63.2} \\ 
        
    \bottomrule
    \end{tabular}
    \label{tab:exp_on_ave_dataset}
\end{table}

\section{Class-wise Performance of the Proposed Two Baselines}
\begin{figure*}[t]
  \centering
\includegraphics[width=0.8\textwidth]{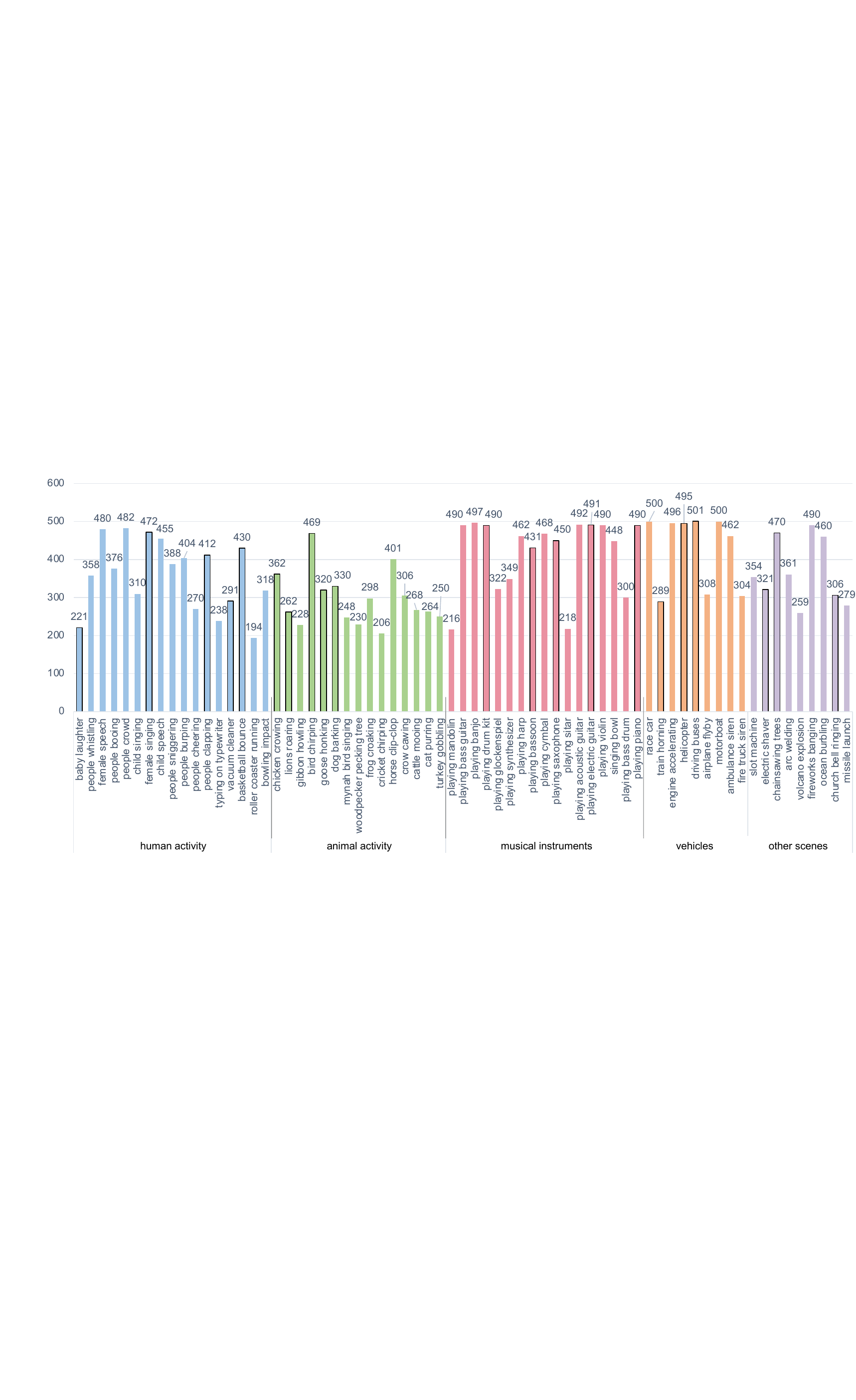}
   \caption{\textbf{Detailed performance of the proposed two baselines on each event class.} 
   }
   \label{fig:class_wise_performance}
\end{figure*}

In Table 2 of our main paper, we present the overall performance of the proposed training-free and fine-tuning baselines on the test set.
Here, we further report their performance on each individual event class.
As shown in Fig.~\ref{fig:class_wise_performance}, the fine-tuning baseline outperforms the training-free baseline in most event classes (approximately 56 out of 67) across all evaluation metrics.
This highlights the benefits of additional fine-tuning on training data.
Moreover, we observe that some event classes, such as \textit{slot machine} and  \textit{chicken crowing}, remain challenging for prediction, suggesting avenues for further improvement in future work.

\section{More Details on Prompts for adapting Video-LLaMA2 to our OV-AVEL task}
In Table 9 of our main paper, we compare the training-free baseline with an advanced audio-visual LLM, namely Video-LLaMA2~\cite{cheng2024videollama}.
Video-LLaMA2 can process video frames and, more importantly, it can handle \textit{general} audio signals that are not limited to human speech, unlike other audio-visual LLMs~\cite{munasinghe2023pg}.
This makes it particularly suitable for the studied OV-AVEL task.
Here, we provide more details on the prompts for adapting Video-LLaMA2 for the OV-AVEL task.
Specifically, we tried several prompts and found the following prompt to be the most robust and effective for making predictions:
``\textit{Instruction: For the given 10-second video, divide it into 10 one-second segments. For each segment, if its audio and visual streams describe the same event, assign the label ``x" as ``1"; otherwise, label this segment as ``0". 
User request: After processing all 10 video segments, you will obtain a list with 10 elements, each element being either ``1" or ``0" according to the above Instruction.
Finally, return the most relevant event category of the video from the candidate category list:
[``airplane flyby", ``ambulance siren", ``arc welding", ``baby laughter", ``basketball bounce", ``bird chirping", ``bowling impact", ``cat purring", ``cattle mooing", ``chainsawing trees", ``chicken crowing", ...(notebly, all event category texts should be listed; here, we omit the remaining ones for simplicity )].
The output format should be: 
``ave:" A python list [x, x, x, x, x, x, x, x, x, x] (replace ``x" with ``1" or ``0" according to the prediction); Insert a line break. ``class:" the most highly relevant class from the given category list (no punctuation needed at the end)}.''
Readers may directly test this prompt on the official demo website using Hugging Face platform provided by authors of Video-LLaMA2~\cite{cheng2024videollama}: \url{https://huggingface.co/spaces/lixin4ever/VideoLLaMA2}.
In this way, we can obtain the audio-visual event predictions of each test video and compare its performance with the proposed training-free baseline, as reported in Table 9 of our main paper.
Additionally, we display some qualitative results in Fig.~\ref{fig:qualitative_1} and Fig.~\ref{fig:qualitative_2} and provide more discussions in Sec.~\ref{sec:qualitative}.

\section{Qualitative Results}\label{sec:qualitative}
\begin{figure*}[t]
  \centering
\includegraphics[width=\textwidth]{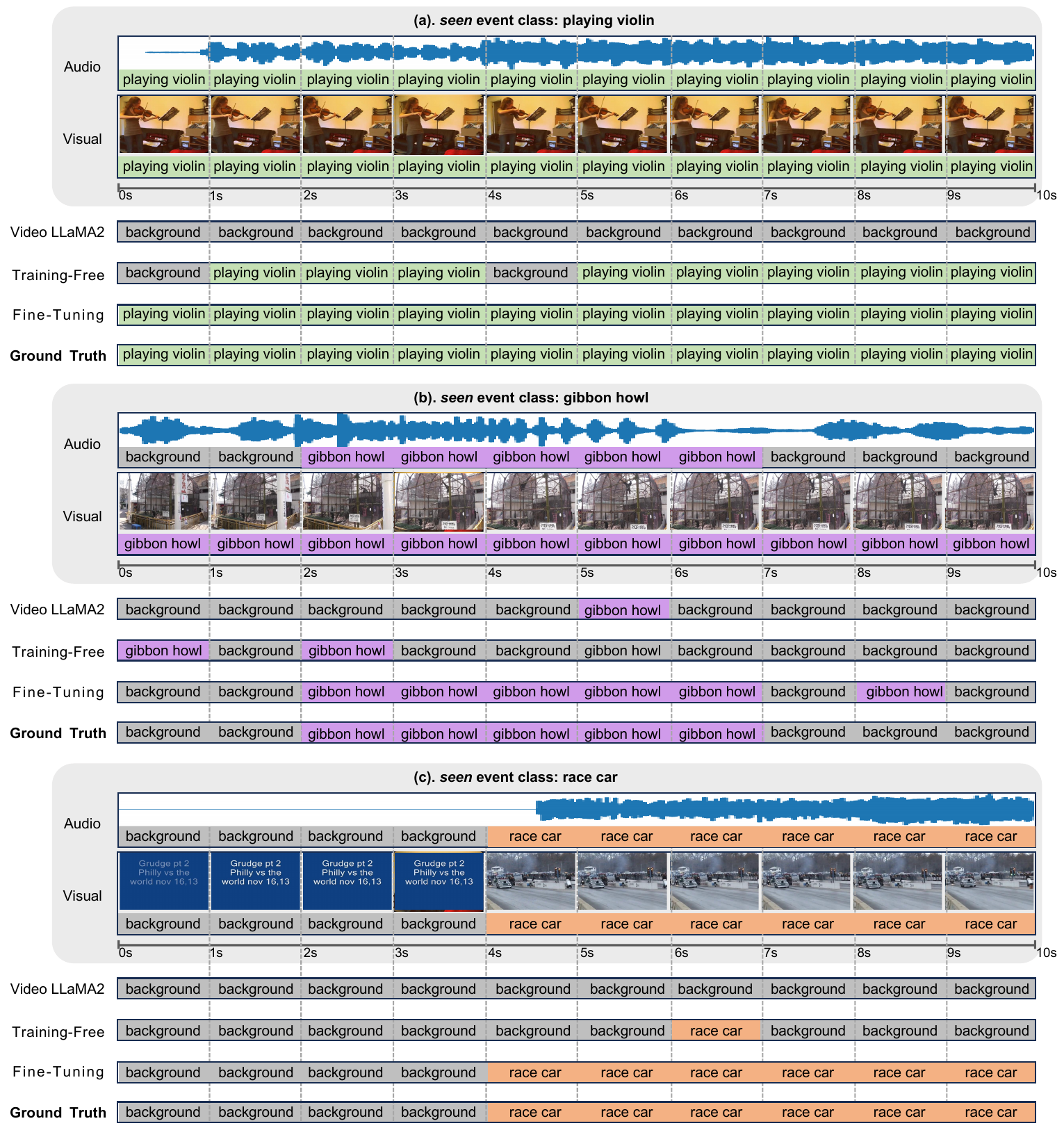}
   \caption{\textbf{Qualitative examples for seen audio-visual event localization. } 
   }
   \label{fig:qualitative_1}
\end{figure*}

\begin{figure*}[t]
  \centering
\includegraphics[width=\textwidth]{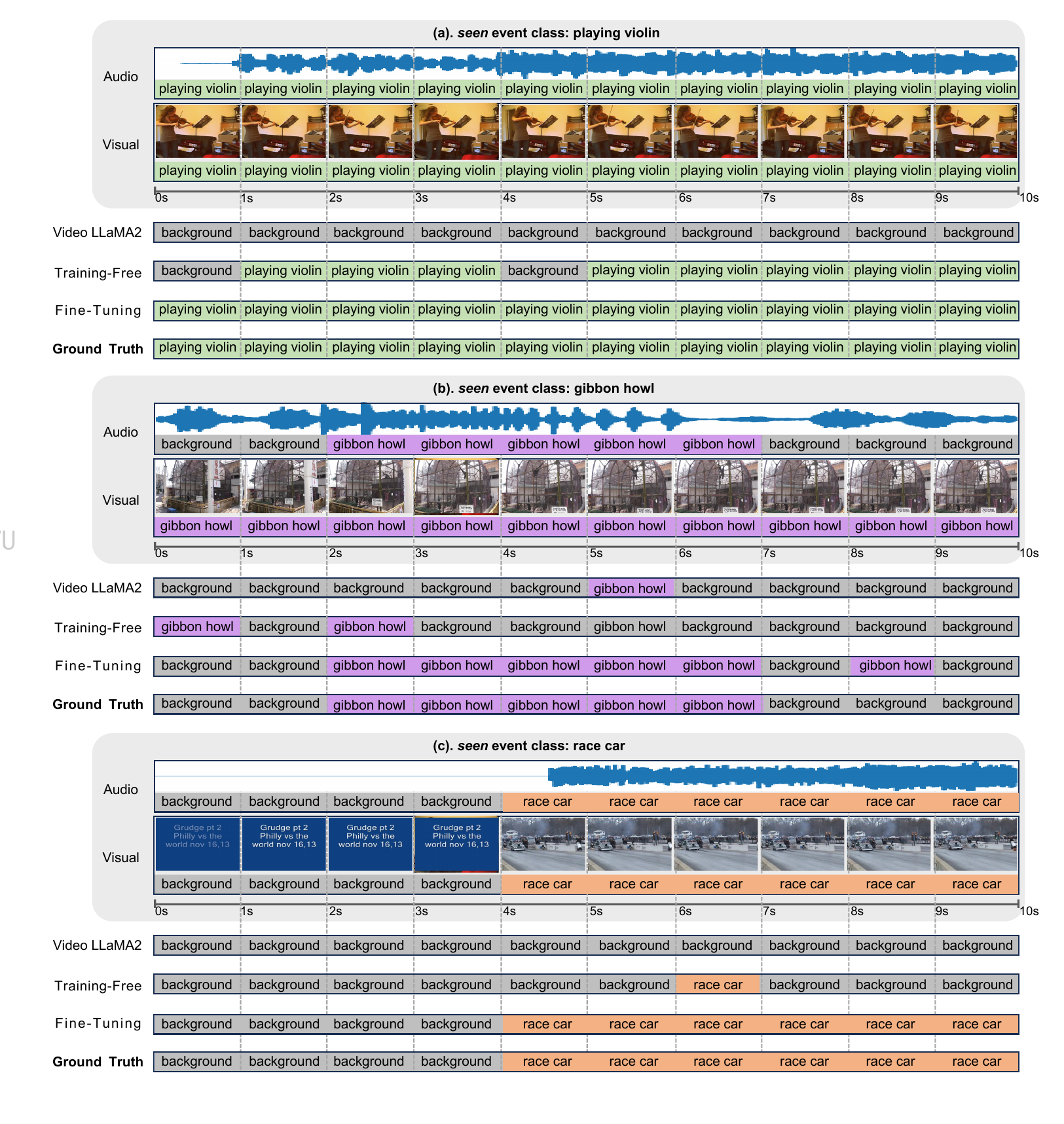}
   \caption{\textbf{Qualitative examples for unseen audio-visual event localization.} 
   }
   \label{fig:qualitative_2}
\end{figure*}

We finally present some intuitive video examples for OV-AVEL, as shown in Fig.~\ref{fig:qualitative_1} and Fig.~\ref{fig:qualitative_2}.
Specifically, we visualize the predictions generated by Video-LLaMA2~\cite{cheng2024videollama}, along with the proposed training-free and fine-tuning baselines.
As shown in the figures, the proposed fine-tuning baseline generally yields more accurate temporal localization results for both seen and unseen events/videos.
For instance, in the three examples shown in Fig.~\ref{fig:qualitative_1}, Video-LLaMA2 tends to predict most video segments as \textit{background}, indicating its limitation in accurately perceiving the audio-visual correspondence at a fine-grained temporal-level.
Although the training-free baseline performs better than Video-LLaMA2, the predictions for some video segments remain unsatisfactory.
In contrast, the fine-tuning baseline performs better in localizing temporal segments containing audio-visual events and classifying the event categories. 
Similar phenomena can also be observed from Fig.~\ref{fig:qualitative_2}.
These qualitative results, along with the quantitative results presented in our main paper, suggest the effectiveness and superiority of the proposed fine-tuning baseline.

\end{document}